\documentclass{article}

\usepackage[final]{neurips_2024}

\usepackage{natbib}
\setcitestyle{numbers,square}
\usepackage{appendix}
\usepackage[utf8]{inputenc} 
\usepackage[T1]{fontenc}    
\usepackage{url}            
\usepackage{booktabs}       
\usepackage{amsfonts}       
\usepackage{nicefrac}       
\usepackage{microtype}      
\usepackage{xcolor}         
\usepackage{multirow}
\usepackage{amsmath}
\usepackage{xspace}
\usepackage{float}
\usepackage[
colorlinks=true,
citecolor=cyan,             
linkcolor=red,              
urlcolor=magenta,           
filecolor=magenta,          
]{hyperref}

\usepackage{graphicx}
\usepackage{caption}
\usepackage{subcaption}
\usepackage{wrapfig}
\usepackage{bbding}
\usepackage{latexsym}
\usepackage{makecell}
\usepackage{colortbl}
\usepackage{arydshln}
\usepackage{bm}
\newcommand{\zhushi}[1]{}

\newcommand{\lmatch}[1]{{\cal L}_{\rm match}(#1)}
\definecolor{MyCyan}{RGB}{224,255,255}

\title{SimVG: A Simple Framework for Visual Grounding with Decoupled Multi-modal Fusion}

\author{
    Ming Dai$^{1}$, Lingfeng Yang$^{2}$, Yihao Xu$^{1}$, Zhenhua Feng$^{3}$, Wankou Yang$^{1,4}$\thanks{Corresponding authors.} \\
  \\
  $^{1}$Southeast University \ \ 
  $^{2}$Nanjing University of Science and Technology \\
  $^{3}$Jiangnan University \ \ 
  $^{4}$Advanced Ocean Institute of Southeast Univerisity, Nantong\\
  \texttt{\footnotesize \{mingdai, 220211848, wkyang\}@seu.edu.cn}, \\ 
  \texttt{\footnotesize yanglfnjust@njust.edu.cn}, \ 
  \texttt{\footnotesize fengzhenhua@jiangnan.edu.cn} \\
}

\begin{document}

	\maketitle
	
	\begin{abstract}
Visual grounding is a common vision task that involves grounding descriptive sentences to the corresponding regions of an image.
Most existing methods use independent image-text encoding and apply complex hand-crafted modules or encoder-decoder architectures for modal interaction and query reasoning.
However, their performance significantly drops when dealing with complex textual expressions.
This is because the former paradigm only utilizes limited downstream data to fit the multi-modal feature fusion. Therefore, it is only effective when the textual expressions are relatively simple.
In contrast, given the wide diversity of textual expressions and the uniqueness of downstream training data, the existing fusion module, which extracts multimodal content from a visual-linguistic context, has not been fully investigated.
In this paper, we present a simple yet robust transformer-based framework, SimVG, for visual grounding.
Specifically, we decouple visual-linguistic feature fusion from downstream tasks by leveraging existing multimodal pre-trained models and incorporating additional object tokens to facilitate deep integration of downstream and pre-training tasks.
Furthermore, we design a dynamic weight-balance distillation method in the multi-branch synchronous learning process to enhance the representation capability of the simpler branch. 
This branch only consists of a lightweight MLP, which simplifies the structure and improves reasoning speed.
Experiments on six widely used VG datasets, \textit{i.e.}, RefCOCO/+/g, ReferIt, Flickr30K, and GRefCOCO, demonstrate the superiority of SimVG.
Finally, the proposed method not only achieves improvements in efficiency and convergence speed but also attains new state-of-the-art performance on these benchmarks.
Codes and models are available at \url{https://github.com/Dmmm1997/SimVG}.
	\end{abstract}

	\section{Introduction}
	\label{sect:introduction}

        \begin{figure*}[t]
		\centering
		\includegraphics[width=1.0\textwidth]{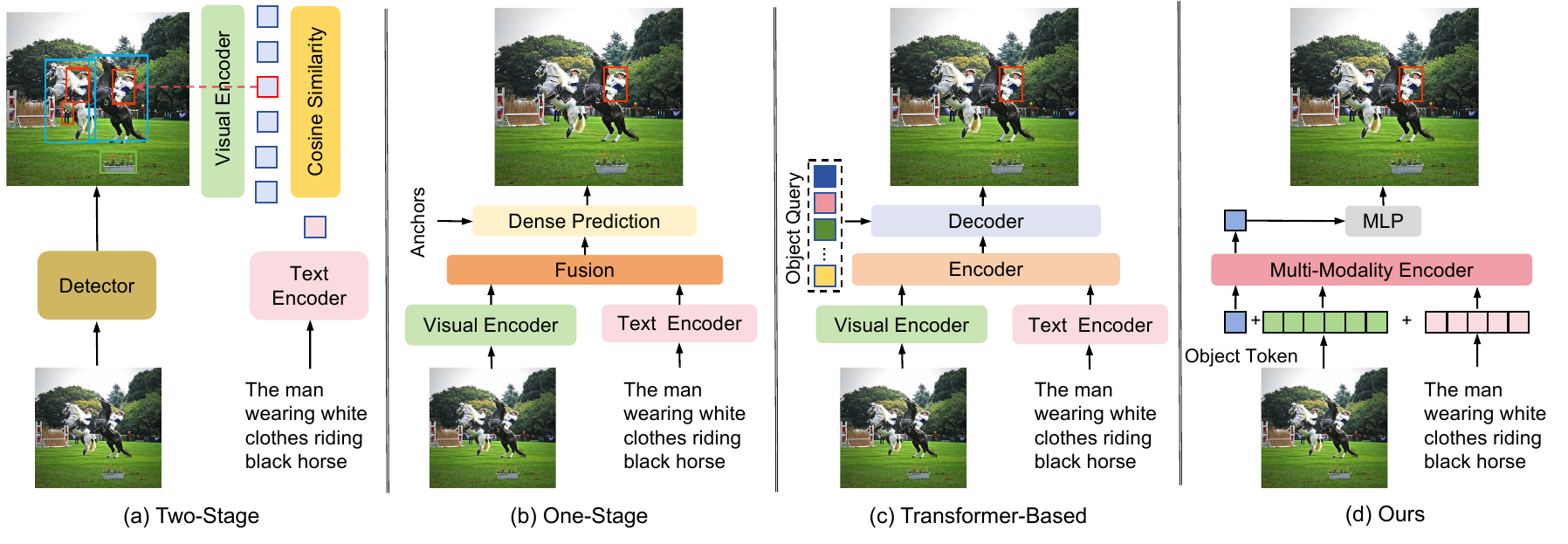}
		\caption{
			An overview of visual grounding structures:
			(a) Two-Stage: Applying a detector for proposals, followed by image-text encoding and feature similarity calculation for region matching.
			(b) One-Stage: Grounding in the fused features through dense prediction.
			(c) Transformer-based: Employing an encoder-decoder structure in the head.
			(d) Proposed SimVG: Utilizing Multi-Modality Encoder for multimodal interaction among object, image, and text tokens, directly applies a lightweight MLP for grounding.
		}
		\label{fig:overview}
        \vspace{-20pt}
	\end{figure*}
	
	Visual grounding (VG) aims to predict the corresponding regions of an image through linguistic expressions. The task necessitates a comprehensive understanding of each modality, as well as the modeling of consistency between image context and text. Some benchmarks focus on addressing \textit{phrase localization}~\cite{referitgame, flickrentities}, which entails locating all objects mentioned in a sentence within an image. Another aspect emphasizes resolving \textit{referring expression comprehension} (REC)~\cite{refcoco/+, refcocogumd, refcocoggoogle}, characterized by only one target corresponding to a sentence. Recently, a new type of \textit{general referring expression comprehension} (GREC)~\cite{GRES, GREC} task has emerged. GREC is similar to REC, but in which a sentence can have multiple targets or no target at all.
	
	Existing visual grounding models can be roughly divided into three categories: two-stage, one-stage, and transformer-based.
	Among them, as shown in Fig.~\ref{fig:overview}(a), two-stage methods~\cite{cmn,vc,mattnet,cmatterase,nmtree} require a pre-trained detector to generate proposals and perform localization through region-text retrieval. These methods rely on a complex module with manually designed mechanisms to achieve query reasoning and multi-modal fusion.
	One-stage methods~\cite{faoa,rccf,mcn,resc}, on the other hand, employ an end-to-end architecture, as shown in Fig.~\ref{fig:overview}(b). Most of them primarily perform dense prediction on multimodal fusion features defined in the form of anchors.
    Some recent algorithms~\cite{transvg, mdetr, seqtr, dynamicmdetr}, depicted in Fig.~\ref{fig:overview}(c), adopt an encoder-decoder architecture to perform multimodal fusion in the encoder and then decode the response target position using an object query similar to DETR\cite{detr}.
    The existing methods share a commonality: they adopt architectures that independently encode each modality before merging them, with multimodal fusion intricately linked to each visual grounding task.
The feature extraction part of these methods generally employs specific classification~\cite{imagenet, imagenet21k} or autoregressive~\cite{bert, alberf} tasks in each modality for pre-training. However, the alignment and mutual understanding between modalities only utilize a limited amount of downstream data, which undoubtedly underestimates the difficulty of achieving mutual understanding between modalities. 
Another observed trend, as noted in~\cite{mdetr, seqtr}, is the notable enhancement in the performance of visual grounding with a significant augmentation of pretraining data on large corpora. This implicitly suggests that leveraging a small amount of downstream data does not fully capitalize on the potential for mutual understanding between images and text.
Nevertheless, this type of pretraining undoubtedly increases the burden of training resources.

        \begin{wrapfigure}{r}{0.5\textwidth}
			\vspace{-20pt}
			\setlength{\fboxrule}{0pt}
			\centering
			\fbox{\includegraphics[width=0.48\textwidth]{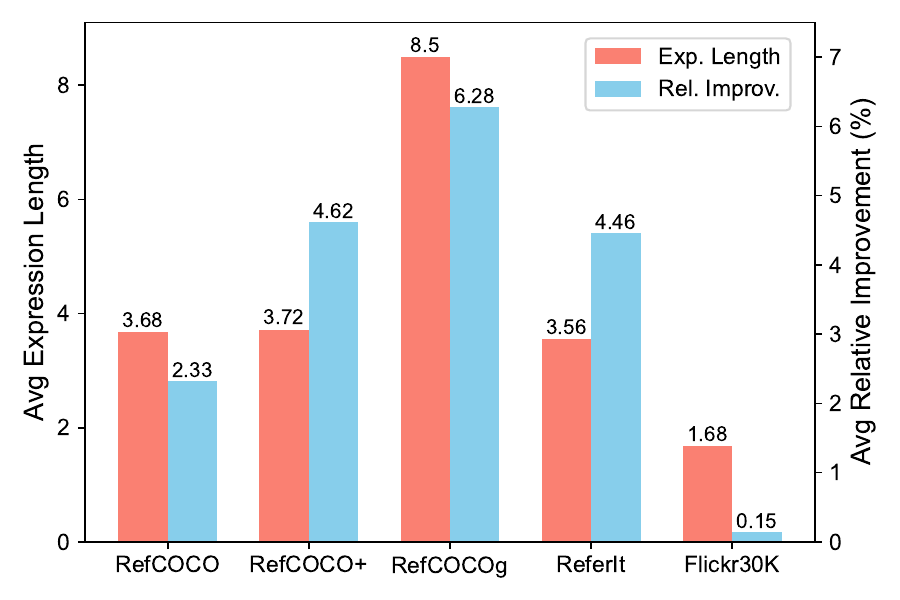}}
			\vspace{-10pt}
			\caption{
				The expression length and relative improvement between Dynamic MDETR~\cite{dynamicmdetr} and SimVG.
			}
			\label{fig_numtoken_improvement}
			\vspace{-10pt}
	\end{wrapfigure}
	
    Furthermore, the mutual understanding between multiple modalities is crucial for downstream tasks. As shown in Fig.~\ref{fig_numtoken_improvement}, for a dataset with long sentence characteristics (like RefCOCOg~\cite{refcocogumd}), adopting the decoupled multimodal understanding method can significantly improve model performance, while the improvement is relatively modest on datasets with short sentences. This observation aligns with our expectation that shorter captions pose less challenge for inter-modal understanding, whereas SimVG's decoupling approach proves advantageous for challenging longer descriptions that require intricate multi-modal comprehension.

    To be specific, we delve into existing multimodal understanding methods in the form of multimodal pre-training. Existing approaches can be broadly categorized into three categories.
    Dual-stream structures~\cite{CLIP,align,declip} encode image-text modalities independently and supervise them with contrastive learning. One-stream models~\cite{UNITER,VILT,alberf,soho} concatenate multimodal features for feature extraction. Other works~\cite{BLIP, vlmo, flava, beit3} use a dual-stream design with a fusion encoder to balance complexity and computational cost, fusing multimodal features in intermediate layers.
	Some methods also have applied multimodal pre-training to VG tasks, such as Dynamic MDETR~\cite{dynamicmdetr}, which uses an image-text encoder pretrained on CLIP~\cite{CLIP} to enhance the performance. Recently, works like CPT~\cite{cpt}, ReCLIP~\cite{reclip}, and FGVP~\cite{fgvp} have improved the performance of zero-shot visual grounding using dual-stream pre-trained models that employ a two-stage-like architecture and prompt engineering techniques. However, these approaches focus on multimodal alignment rather than mutual understanding. Thus, the dual-stream with fusion encoder architecture~\cite{BLIP, vlmo, flava, beit3} has been thrown into our sight. Specifically, based on BEiT-3~\cite{beit3}, we propose a simple framework called SimVG that decouples multimodal fusion from downstream tasks and simultaneously encodes image, text, and object tokens, as illustrated in Fig.~\ref{fig:overview} (d).
The decoder structure used for query reasoning, similar to DETR~\cite{detr}, is effective but inevitably increases the model's complexity and computational overhead.
We aim to develop a more efficient and simpler architecture for visual grounding. To achieve this, in addition to the decoder branch, we introduce a lightweight token branch. This branch leverages object tokens that are deeply integrated with image-text features to enhance grounding performance. To ensure that the token branch maintains high performance while enabling efficient inference, we adhere to the principles of the existing knowledge distillation methods~\cite{huang2023teach, chen2022d, chang2023detrdistill} and introduce an innovative dynamic weight-balance distillation (DWBD). This approach enhances the token branch's capability by dynamically assigning weights to the decoder's predictions and ground truth at various learning stages, facilitating more effective learning in the token branch.
Furthermore, we introduce a text-guided query generation (TQG) module to integrate textual information into queries, enabling the adaptive incorporation of textual prior knowledge into object queries. Notably, the design of the TQG module allows for the expansion of object tokens, increasing the number of objects it can handle and thereby adapting effectively to the GREC~\cite{GREC} task.
The experiments demonstrate that, by decoupling multimodal fusion from downstream tasks, SimVG achieves rapid convergence and superior performance even with a limited amount of data. Coupled with our proposed DWBD and TQG modules, SimVG sets new state-of-the-art performance across various benchmarks.
 
	Our main contributions are summarized as follows:
	
	{
		$\bullet$ We propose a simple yet strong visual grounding architecture called SimVG, which decouples multimodal understanding from downstream tasks and fully utilizes the inter-modal fusion ability of existing multimodal pre-trained models. To the best of our knowledge, SimVG is the first to employ a unified structure for encoding image, text, and object tokens in visual grounding.
	}
	
	{
		$\bullet$ We propose a novel dynamic weight-balance distillation (DWBD) to dynamically allocate weights to decoder predictions and ground truth at different stages of multi-branch synchronous learning, aiming to minimize the discrepancy between the token and decoder branches. Furthermore, we introduce a text-guided query generation (TQG) module to incorporate textual prior information into object queries, thereby extending its applicability to the GREC task.
	}
	
	{
		$\bullet$ The proposed SimVG architecture has demonstrated state-of-the-art performance across six prominent datasets, while also exhibiting notable gains in efficiency and convergence speed. Particularly noteworthy is that SimVG (ViT-B/32) achieves these results with just 12 hours of training on a single RTX 3090 GPU when applied to the RefCOCO/+/g datasets.
	}

	\section{Related Work}

	\paragraph{Vision-Language Pre-Training.}
	{
Existing vision-language pretraining (VLP) models can be broadly categorized into three main types: one-stream, dual-stream, and dual-stream with fusion encoder architectures. 
  One-stream models~\cite{UNITER,VILT,alberf,soho} process both image and text inputs in a single stream, concatenate image and text embeddings, and interact cross-modals information throughout the whole feature extraction process.
  In contrast, dual-stream models~\cite{CLIP,align,declip} employ separate encoders for each modality. These models do not concatenate modalities at the input level. Instead, the interaction between pooled image and text vectors occurs at a shallow layer.
  Dual-stream models with fusion encoder~\cite{BLIP, vlmo, flava, beit3} combine aspects of both one-stream and dual-stream models. They allow for intermediate interaction between modalities, potentially offering a balance between complexity and performance.
  In this paper, we improve the performance of visual grounding by decoupling multi-modal fusion from downstream tasks into upstream VLP models~\cite{beit3}.
	}

	\paragraph{Referring Expression Comprehension.}
	{
		Early approaches in REC typically followed a two-stage pipeline~\cite{cmn,vc,parallelattention,mattnet,cmatterase,dga,rvgtree,nmtree}. This pipeline involves first extracting region proposals~\cite{fasterrcnn}, which are then ranked based on their similarity scores with the language query.
		In contrast, a more recent line of research~\cite{realgin,faoa,rccf,mcn,resc,lbyl} advocates for a simpler and faster one-stage pipeline based on dense anchors.
		Several recent approaches~\cite{transvg,trar,mdetr,vlt,reftr, seqtr} have employed a transformer-based structure~\cite{transformer} for multi-modal fusion. 
		Furthermore, with the vigorous development of multimodal large language models (MLLM)~\cite{mllmsurvey}, some of the latest methods have further enhanced the generalization performance of REC through zero-shot~\cite{fgvp, reclip} or fine-tuning~\cite{uninext, kosmos} methods, leveraging the powerful capabilities of large models~\cite{llava} and general models~\cite{sam}.
		In contrast to existing methods, our proposed SimVG method directly feeds object, image, and text tokens into the multi-modality encoder for multimodal feature interaction. We eschew the complex encoder-decoder structure and perform visual grounding directly using a simple MLP.
	}
	
	\paragraph{Knowledge Distillation in Object Detection.}
	{
		The majority of research in knowledge distillation has primarily focused on classification tasks~\cite{guan2020differentiable,xie2020self,yuan2020revisiting}. Several studies~\cite{liu2019structured,wang2020intra,zhang2020improve,wang2019distilling} have extended knowledge distillation techniques to dense prediction tasks, such as semantic segmentation and object detection. These works commonly exploit pixel-wise correlations or channel-wise interactions between dense features of teacher and student models. Recently, there has been a growing interest in developing tailored knowledge distillation losses for query-based detectors like DETR~\cite{detr}, as demonstrated by works such as~\cite{huang2023teach,chen2022d,chang2023detrdistill}.
        Unlike previous methods that guide a lightweight student model using a pre-trained teacher model, this paper introduces knowledge distillation during synchronous learning to enhance the performance of the lightweight branch.
	}
	
	\begin{figure*}[t]
		\centering
		\includegraphics[width=1.0\textwidth]{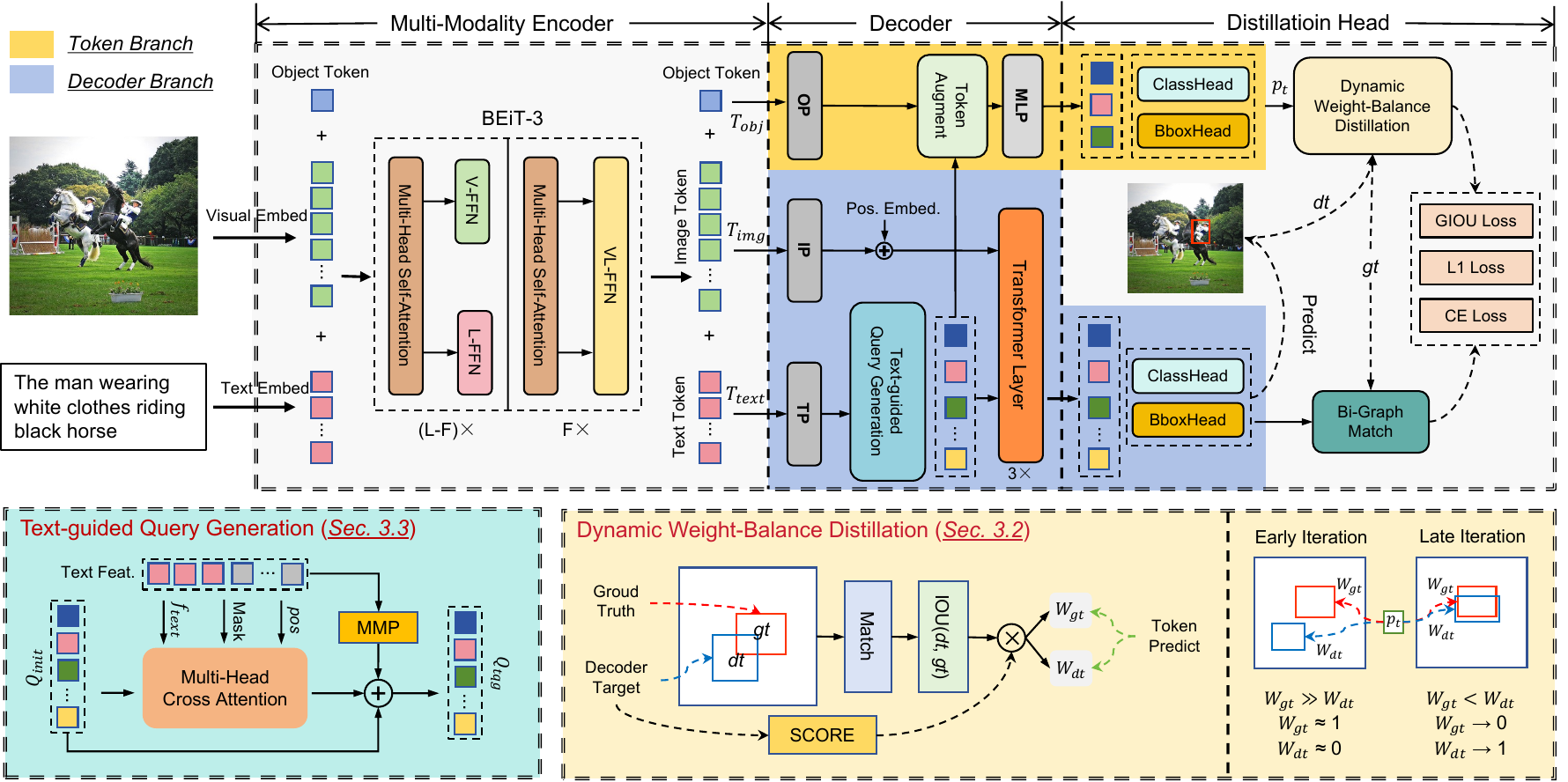}
		\caption{Overview of the proposed SimVG. The token branch refers to the upper light yellow region, while the decoder branch refers to the lower light blue region. During model inference, we can independently apply the more lightweight token branch to improve inference speed and simplify the model architecture.}
		\label{fig:model}
            \vspace{-10pt}
	\end{figure*}

	\section{The Proposed Method}
 \label{methods}
	
	\subsection{The Overview of SimVG}
	\label{sect:The Overview of SimVG}
	As shown in Fig.~\ref{fig:model}, the SimVG structure can be roughly divided into three parts: multi-modality encoder, decoder, and head. The multi-modality encoder adopts a structure similar to BEiT-3~\cite{beit3}, and additionally sets a learnable object token. The decoder is divided into two branches: one is similar to the transformer decoder in DETR~\cite{detr} (\textit{\underline{decoder branch}}), and the other utilizes a lightweight MLP (\textit{\underline{token branch}}). The head is referred to as the "Distillation Head". Unlike conventional prediction heads, to reduce the performance gap between the token and decoder branches, we employ a dynamic weight-balance distillation (DWBD) to minimize the performance difference between the two branches during synchronous learning. 
	
	\noindent{\bf Multi-Modalilty Encoder.}
    The input of SimVG include an image ${I}\in\mathbb{R}^{3\times{H}\times {W}}$ and a caption text ${T}\in\Omega^{M}=\{t^1,t^2,...,t^M\}$, where $\Omega$ denotes the set of words.
	First, the image is compressed to a scale of 1/32 of the original size using visual embedding to obtain ${P}_{img}=\{p^1,p^2,...,p^N\}$. The text is then mapped to the vector space ${L}_{text}=\{l^1,l^2,...,l^K\}$ and an text padding mask ${L}_{mask}=\{l^1_m,l^2_m,...,l^K_m\}$. 
	Additionally, we define a learnable object token $T_{obj}$ as the target feature for the token branch. The query and attention padding mask of the transformer can be generated as:
	\begin{equation}
		\setlength{\abovedisplayskip}{2pt}
		\setlength{\belowdisplayskip}{2pt}
		\begin{aligned}
			\text{Query} = \{T_{obj},p^1,...,p^N,l^1,...,l^K\}, \ \ \ \text{Mask} = \{{T_{obj}}_m,p^1_m,...,p^N_m,l^1_m,...,l^K_m\}.
		\end{aligned}
		\vspace{0pt}
		\label{eq_token_mask}
	\end{equation}
	In the case where the image has no padding, ${T_{obj}}_m$ and $\{p^1_m,p^2_m,...,p^N\}$ are set to 0. In the independent encoding part, FFN adopts a setting of non-shared weights between the image and text modalities, while the rest remains largely the same with the original ViT~\cite{vit} model.
	
	\noindent{\bf The Decoder Branch.} We first map the channel dimensions of the image tokens $T_{img} \in{\mathbb{R}^{\frac{H}{32}\times \frac{H}{32} \times C}}$ and text tokens $T_{text} \in{\mathbb{R}^{N_{text} \times C}}$ using a linear layer without sharing weights. Then, on the text side, we apply the text-guided query generation ($\text{TQG}$) module to interact the predefined object queries $Q_{obj} \in{\mathbb{R}^{N_{oq} \times C}}$ with the text tokens. On the image side, additional positional encoding is applied. Lastly, cross-attention interaction is performed through transformer modules. The entire process can be represented as follows:
	\begin{equation}
		\setlength{\abovedisplayskip}{2pt}
		\setlength{\belowdisplayskip}{2pt}
		\begin{aligned}
			Q_{decoder}=\mathbf{MCA}(\mathbf{IP}(T_{img})+{pos}, \mathbf{TQG}(\mathbf{TP}(T_{text}), Q_{obj})),
		\end{aligned}
		\vspace{0pt}
		\label{eq_token_branch}
	\end{equation}
	where $\mathbf{TP}$ and $\mathbf{IP}$ refer to the text projection and image projection. $\mathbf{MCA}$ refers to multi-head cross attention. $pos$ refers to position encoding, which applies 2D absolute sine encoding.
	
	\noindent{\bf The Token Branch.} We employ a linear layer $\mathbf{OP}$ to project object token $T_{obj} \in{\mathbb{R}^{1 \times C}}$ and use the results of TQG to augment the object token. Lastly, we use an MLP layer to further interact with and enrich the representation of the object token. The process of this branch can be defined as:
	\begin{equation}
		\setlength{\abovedisplayskip}{2pt}
		\setlength{\belowdisplayskip}{2pt}
		\begin{aligned}
			Q_{token}=\mathbf{MLP}(\mathbf{OP}(T_{obj})+\mathbf{TQG}(\mathbf{TP}(T_{text}),Q_{obj})).
		\end{aligned}
		\vspace{0pt}
		\label{eq_token_branch2}
	\end{equation}
	\noindent{\bf The Distillation Head.} We adopt the same Hungarian matching as DETR~\cite{detr} for the decoder branch. The matching cost consists of three parts: binary cross-entropy loss, l1 loss, and giou loss~\cite{giou}. However, to further simplify the inference pipeline, we use the decoder branch as a teacher to guide the learning of the token branch during the whole training process. Therefore, the complete loss can be represented as follows:
	\begin{equation}
		\setlength{\abovedisplayskip}{2pt}
		\setlength{\belowdisplayskip}{2pt}
		\begin{aligned}
			{\cal L}_{total} =  {{\cal L}_{det}(p_{d}, gt)} +  {{\cal L}_{dwbd}}, \ \ \ \  {{\cal L}_{det}} = \lambda_1{{\cal L}_{ce}}+\lambda_2{\cal L}_{l1}+\lambda_3{{\cal L}_{giou}},
		\end{aligned}
		\vspace{0pt}
		\label{eq_total_loss}
	\end{equation}
	where $\lambda_1$, $\lambda_2$, and $\lambda_3$ are set to 1, 5, and 2. $p_{d}$ refers to the decoder branch prediction. ${{\cal L}_{dwbd}}$ refers to the dynamic weight-balance distillation loss, which will be explained in the next part.

	
	\subsection{Dynamic Weight-Balance Distillation}
	\label{sect:Dynamic Weight-Balance Distillation}
    To make SimVG both efficient and effective, knowledge distillation is introduced, which leverages the predictions from the decoder branch as a teacher to guide the predictions of the token branch. Since the two branches share the features of the multi-modality encoder, training the teacher model independently using traditional knowledge distillation methods is not feasible. Instead, we employ a synchronous learning approach for both branches. This approach requires a delicate balance, ensuring that the performance of the teacher model is not compromised while maximizing the transfer of knowledge from the teacher branch to the student branch.
	
	Therefore, we design a dynamic weight-balance distillation (DWBD), whose architecture is shown in Fig.~\ref{fig:model}.
	Let us denote the ground truth by $y$, and the set of $N_q$ decoder prediction by $\hat{y}^d = \{\hat{y}_i^d\}_{i=1}^{N_{q}}$.
	To find a bipartite matching between these two sets, we search for a permutation of $N_q$ elements $\sigma \in \Sigma_{N_q}$ with the lowest cost:
	\begin{equation}
            \setlength{\abovedisplayskip}{2pt}
		\setlength{\belowdisplayskip}{2pt}
		\label{eq:matching}
		\hat{\sigma} = \textbf{ARGMIN}_{\sigma\in\Sigma_{N_q}} \sum_{i}^{N_{q}} \lmatch{y_i, \hat{y}_{\sigma(i)}},
	\end{equation}
	where $\lmatch{y_i,\hat{y}_{\sigma(i)}}$ is a pair-wise matching cost between the ground truth $y_i$ and a prediction with index $\sigma(i)$. 
	After pairing, the next step is to assess the decoder branch's understanding capability at the current stage. This is done by measuring the confidence of the joint target box of the decoder target $dt$ and ground truth $gt$ based on the IOU of their pairing:
	\begin{equation}
            \setlength{\abovedisplayskip}{2pt}
		\setlength{\belowdisplayskip}{2pt}
		\label{eq:weight}
		W_{dt} = \frac{1}{N_{gt}}\sum_{i}^{N_{gt}}\left[\textbf{IOU}(b_{i}, \hat{b}_{\hat{\sigma}}(i)) \times \textbf{SCORE}(\hat{p}_{\hat{\sigma}(i)})\right],
	\end{equation}
	where $N_{gt}$ is the number of ground truth boxes, $\text{SCORE}$ represents the foreground score extracted from the predictions. $W_{dt}$ can be seen as a reflection of the current stage's decoder branch capability, where a higher value indicates a stronger confidence. Lastly, ${\cal L}_{dwbd}$ can be expressed as follows:
	\begin{equation}
            \setlength{\abovedisplayskip}{2pt}
		\setlength{\belowdisplayskip}{2pt}
		\label{eq:dwbd}
		{\cal L}_{dwbd}= \gamma_1(W_{dt} \times {\cal L}_{det}(p_t, dt)) + \gamma_2(W_{gt} \times {\cal L}_{det}(p_t, gt)), \ \ \ \ W_{gt}=1-W_{dt},
	\end{equation}
	where ${\cal L}_{det}$ is computed exactly the same as in Eq.~\ref{eq_total_loss}. $p_t$ refers to the token branch prediction. $ \gamma_1$ and $\gamma_2$ are set to 2 and 1 in this paper.
	By design, in the early stage of network training, $W_{gt} \gg W_{dt}$, the training process of the entire token branch is guided by the ground truth. However, in the later training stage, $W_{gt} < W_{dt}$, the guidance from the decoder target becomes more significant. This dynamic adjustment of weights during training is the core idea of the proposed DWBD. 
        We will further analyze the changes in \(W_{dt}\) and \({\cal L}_{dwbd}\) in Sec.~\ref{ab:distill}.
	Additionally, to further enhance the performance of the token branch, we use a two-stage distillation approach. In the first stage, we train the decoder branch separately. In the second stage, we apply DWBD under the premise of synchronous learning for the two branches.
	
	\subsection{Text-guided Query Generation}
	\label{sect:Text-guided Query Generation}
	The initial object queries, $Q_{init}$, are defined using learnable embeddings, without any prior information to guide them. From a macro perspective, visual grounding involves using text as a query to locate optimal regions in an image. Embedding text into queries to adaptively provide priors offers a viable solution. Therefore, we propose a text-guided query generation (TQG) module to generate object queries with text priors. As illustrated in Fig.~\ref{fig:model}, the process of generating queries through TQG can be expressed as follows:
	\begin{equation}
		\label{eq:tqg}
            \setlength{\abovedisplayskip}{2pt}
		\setlength{\belowdisplayskip}{2pt}
		Q_{tqg} = \textbf{MCA}(Q_{init}, f_{text}+pos, \text{Mask}) + \textbf{MMP}(f_{text}, \text{Mask}) + Q_{init},
	\end{equation}
	where $f_{text}\in{\mathbb{R}^{K\times C}}$ is the feature after text projection, \text{Mask} is consistent with Eq.~\ref{eq_token_mask}, and $pos$ here is 1D absolute sine positional encoding. 
	$\textbf{MMP}$ is the process of filtering out valid tokens from $f_{text}$ using the \text{Mask} and applying a max operation: $\textbf{MMP}(f, m) = \max_{i} \left[f_{i} \times (\sim m_i) \right]$.
	
	\section{Experimental Results}
	
	\subsection{Datasets and Evaluation Metric}
        The experiments in this paper are conducted on six widely used datasets: RefCOCO/+/g~\cite{refcoco/+, refcocoggoogle, refcocogumd}, Flickr30K Entities~\cite{flickrentities}, ReferItGame~\cite{referitgame}, and GRefCOCO~\cite{GREC}. For the referring expression comprehension and phrase localization tasks, Precision@0.5 is used as the evaluation metric. The GRefCOCO dataset is evaluated using the Precision@(F$_1$=1, IoU$\ge$0.5) and N-acc metrics. More descriptions about datasets and evaluation metrics are provided in Appendix~\ref{sec:datasets} and ~\ref{sect:evaluation metrics}.


	\subsection{Implementation Details}
 \label{implementation details}
	We train SimVG 30 epochs for REC and phrase localization, and 200 epochs for GREC, using a batch size of 32. 
 Following standard practices, images are resized to 640$\times$640, and the length of language expressions is trimmed to 20 for all the datasets. For pre-training, SimVG is trained for 30 epochs and then fine-tuned for another 10 epochs. 
    The pre-training experiments are run on 8 NVIDIA RTX 3090 GPUs. All the other experiments are conducted on 2 NVIDIA RTX 4090 GPUs. 
    More implementation details are reported in Appendix~\ref{appendix_implementation_details}.
	
	\subsection{Comparison with The State-of-the-art}
	\label{sect:comparisons with state-of-the-art methods}
	In this part, we compare our SimVG with the SOTA methods on six mainstream datasets. We combine the results of RefCOCO/+/g, ReferItGame and Flickr30K datasets in Table~\ref{table:sotaonrec}, and the results of GREC are reported in Table~\ref{table:sotaongrec}. Table~\ref{table:sotaonpretrain} reports the results pre-trained on the large corpus of data.
	
    According to Table~\ref{table:sotaonrec}, our model performs better than two-stage models,  especially MAttNet~\cite{mattnet} while being 7 times faster. We also surpass one-stage models that exploit prior and expert knowledge, with +14\% absolute average improvement over ReSC~\cite{resc}.
	Additionally, the use of a patch stride of 32 and a lightweight head design has enabled SimVG to achieve an inference speed of only 44 ms on a GTX 1080Ti GPU. For transformer-based models, SimVG surpasses the recent SOTA method Dynamic MDETR~\cite{dynamicmdetr} with an average of up to 4.4\% absolute performance improvement. 
	
	As shown in Table~\ref{table:sotaonrec}, on the ReferItGame and Flickr30K Entities datasets which mostly contain short noun phrases, the performance boosts to 74.83 and 82.04 with a large margin over the previous one-stage method~\cite{resc}. Compared to existing transformer-based methods~\cite{transvg, reftr, seqtr}, SimVG still significantly outperforms most SOTA methods by approximately 3.4 points on the ReferItGame dataset, and it also slightly outperforms Dynamic MDETR~\cite{dynamicmdetr} on the Flickr30k dataset. Furthermore, scaling the model from base to large has led to significant improvements across all the datasets.
	
	SimVG can be seamlessly extended to GREC without any network modification. As shown in Table~\ref{table:sotaongrec}, SimVG achieves a significant improvement over existing publicly available methods on the GRefCOCO dataset, with an average increase of 9 points, surpassing UNINEXT~\cite{uninext}. 
	
	Table~\ref{table:sotaonpretrain} demonstrates that when pre-training on a large corpus of image-text pairs, SimVG exhibits greater data efficiency as compared with most of the existing SOTA methods.
	Despite utilizing only 28K images, which is nearly six times fewer than MDETR~\cite{mdetr}, and three times fewer than RefTR~\cite{reftr}, SimVG still achieves SOTA performance, surpassing most existing methods by a significant margin. Compared to MDETR, SimVG demonstrates an average improvement of 5 points, and compared to the recent SOTA model GroundingDINO~\cite{groundingdino}, it achieves an average improvement of 2 points. Moreover, increasing the volume of pre-training data further enhances performance. 
    Additionally, SimVG applies a lighter transformer structure in the head. Specifically, SimVG-TB only uses 1.58 million parameters, which is smaller than some lightweight models~\cite{seqtr, reftr}.
    Lastly, we observe that scaling the multimodal encoder from ViT-B to ViT-L results in the performance of the lightweight token branch surpassing that of its teacher model. We hypothesize that as the model size increases, the reliability of the decoder branch's performance improves, helping to mitigate the impact of mislabeled ground truth data. This, in turn, enhances the generalization ability of the token branch, further demonstrating the effectiveness of the DWBD method.
	
	\begin{table*}[t]
		\setlength{\tabcolsep}{4pt}
		\renewcommand{\arraystretch}{1.1}
            \vspace{-10pt}
		\scriptsize
		\begin{center}
                \caption{Comparison with some SOTA methods. RN101 refers to ResNet101~\cite{resnet}, DN53 denotes DarkNet53~\cite{yolov3}, ViT-B/32 means ViT-Base~\cite{vit} with stride of 32 in visual embedding. $*$ denotes testing with a NVIDIA RTX 3090 GPU, while other entries are tested with a GTX 1080Ti GPU. SimVG-TB and SimVG-DB refer to the SimVG model using only the token and decoder branch for inference, respectively.}
			\begin{tabular}{l|c|ccc|ccc|ccc|c|c|c}
				\hline
				\multirow{2}{*}{Models} & \multirow{2}{*}{\makecell{Visual \\ Encoder}} & \multicolumn{3}{c|}{RefCOCO} & \multicolumn{3}{c|}{RefCOCO+} & \multicolumn{3}{c|}{RefCOCOg} & ReferIt & Flickr30k & Time \\
				& & val & testA & testB & val & testA & testB & val-g & val-u & test-u & test&test &(ms) \\
				
				\hline
				
				\multicolumn{14}{l}{\textbf{Two-Stage}} \\
				
				\hline
				
				MAttNet~\cite{mattnet} & RN101 & 76.40 & 80.43 & 69.28 & 64.93 & 70.26 & 56.00 & - & 66.58 & 67.27 &29.04& -& 320 \\
				CM-Att-Erase~\cite{cmatterase} & RN101 & 78.35 & 83.14 & 71.32 & 68.09 & 73.65 & 58.03 & - & 67.99 & 68.67 &- &- & - \\
				DGA~\cite{dga} & VGG16 & - & 78.42 & 65.53 & - & 69.07 & 51.99 & - & - & 63.28 & - &- & 341 \\
				RvG-Tree~\cite{rvgtree} & RN101 & 75.06 & 78.61 & 69.85 & 63.51 & 67.45 & 56.66 & - & 66.95 & 66.51 & - & - & - \\
				NMTree~\cite{nmtree} & RN101 & 76.41 & 81.21 & 70.09 & 66.46 & 72.02 & 57.52 & 64.62 & 65.87 & 66.44 & - &-  & - \\
				\hline
				\multicolumn{14}{l}{\textbf{One-Stage}} \\
				\hline
				RealGIN~\cite{realgin} & DN53 & 77.25 & 78.70 & 72.10 & 62.78 & 67.17 & 54.21 & - & 62.75 & 62.33 &-& -& 35 \\
				FAOA~\cite{faoa} & DN53 & 71.15 & 74.88 & 66.32 & 56.86 & 61.89 & 49.46 & - & 59.44 & 58.90& 60.67 & 68.71& 39 \\
				RCCF~\cite{rccf} & DLA34 & - & 81.06 & 71.85 & - & 70.35 & 56.32 & - & - & 65.73 &63.79 & - & \textbf{25} \\
				MCN~\cite{mcn} & DN53 & 80.08 & 82.29 & 74.98 & 67.16 & 72.86 & 57.31 & - & 66.46 & 66.01 & -& -& 56 \\
				$\text{ReSC}_{L}$~\cite{resc} & DN53 & 77.63 & 80.45 & 72.30 & 63.59 & 68.36 & 56.81 & 63.12 & 67.30 & 67.20 &64.60 & 69.28& 36 \\
				LBYL~\cite{lbyl} & DN53 & 79.67 & 82.91 & 74.15 & 68.64 & 73.38 & 59.49 & 62.70 & - & - &  67.47  &  - & \underline{30} \\
				\hline
				
				\multicolumn{14}{l}{\textbf{Transformer-Based}} \\
				
				\hline
				TransVG~\cite{transvg} & RN101 & 81.02 & 82.72 & 78.35 & 64.82 & 70.70 & 56.94 & 67.02 & 68.67 & 67.73 & 70.73 & 79.10& 62 \\
				TRAR~\cite{trar} & DN53 & - & 81.40 & 78.60 & - & 69.10 & 56.10 & - & 68.90 & 68.30 &  - & - & - \\
				VGTR~\cite{vgtr}  & RN50 & 78.29 &81.49& 72.38 &63.29 &70.01 &55.64 &61.64 &64.19 &64.01& 63.63 & 75.44 &-\\
				SeqTR~\cite{seqtr} & DN53 & 83.72 & 86.51 & 81.24 & 71.45 & 76.26 & 64.88 & 71.50 & 74.86 & 74.21 &69.66 & 81.23& 50 \\
				VLTVG~\cite{vltvg} & RN50 &  
				84.53 & 87.69 & 79.22 & 73.60 & 78.37 &64.53 &72.53 &74.90&73.88&  71.60& 79.18& 79$^{*}$ \\
				TransCP~\cite{transcp} & RN50 &  
				84.25 & 87.38 & 79.78 & 73.07 & 78.05 &63.35 &72.60 &- &-& 72.05&80.04 & 74$^{*}$ \\
				Dyn.MDETR~\cite{dynamicmdetr} & ViT-B/16 & 85.97 & 88.82& 80.12 &74.83 &81.70 &63.44 &72.21 & 74.14 & 74.49& 70.37& {81.89}& -\\
				\hline
				\rowcolor{gray!8} SimVG-TB~(ours) & ViT-B/32 & 87.07&89.04 & 83.57& {78.84}&{83.64}&70.67& 77.66 & 79.82 & 79.93 &74.59 & 81.59&44  \\
    
				\rowcolor{gray!8} SimVG-DB~(ours)& ViT-B/32 & {87.63}& {90.22}& {84.04} &78.65 & 83.36&{71.82}& {78.81} & {80.37} & {80.51} & {74.83}& {82.04} &52  \\

                \hline
                
                \rowcolor{gray!15} SimVG-TB~(ours)& ViT-L/32 & \textbf{90.61}& \textbf{92.53}& \textbf{87.68} &\textbf{85.36} & \textbf{89.61}&\textbf{79.74}&  \underline{79.34} & \textbf{85.99} & \textbf{86.83}& \textbf{79.30} & \underline{82.61} &  101 \\
                
                \rowcolor{gray!15} SimVG-DB~(ours)& ViT-L/32 & \underline{90.51}& \underline{92.37}& \underline{87.07} &\underline{84.88} & \underline{88.50}&\underline{78.66}&  \textbf{80.42} & \underline{85.72} & \underline{86.70}& \underline{78.75} & \textbf{83.15} & 116 \\
				\hline
			\end{tabular}
                \label{table:sotaonrec}
                \vspace{-10pt}
		\end{center}
	\end{table*}
	
	\begin{table*}[t]
		\setlength{\tabcolsep}{4pt}
		\renewcommand{\arraystretch}{1.1}
		\begin{center}
			\scriptsize
			\centering
                \caption{GREC benchmark results on GRefCOCO dataset. Threshold is set to 0.7 for all the methods.}
                \vspace{-5pt}
			\setlength{\tabcolsep}{0.96mm}{\begin{tabular}{l|cc|cc|cc|cc}
					\hline
					\multirow{2}{*}{Methods} &\multirow{2}{*}{\shortstack{Visual\\Encoder}}  & \multirow{2}{*}{\shortstack{Textual\\Encoder}} & \multicolumn{2}{c|}{val} & \multicolumn{2}{c|}{testA} & \multicolumn{2}{c}{testB} \\
					&  & &Prec@(F$_1$@0.5)   & N-acc.  &Prec@(F$_1$@0.5)   &  N-acc.   &Prec@(F$_1$@0.5)    & N-acc.   \\
					
					\hline
					
					MCN~\cite{mcn}   & DN53 & GRU& 28.0 &30.6  & 32.3 & 32.0 & 26.8 & 30.3 \\
					VLT~\cite{vlt}   & DN53 & GRU & 36.6 & 35.2  & 40.2 & 34.1 & 30.2 & 32.5 \\
					MDETR~\cite{mdetr}  & RN101 & RoBERTa & 42.7 & 36.3 & 50.0 & 34.5 & 36.5 & 31.0   \\
					UNINEXT~\cite{uninext} & RN50 & BERT &58.2  &  50.6 & 46.4 & 49.3 & 42.9 & 48.2 \\
					
					\hline
					
					\rowcolor{gray!8} SimVG-TB~(ours) & ViT-B/32 & / & \underline{61.3} & \textbf{56.1} & \underline{61.7} &\textbf{58.0} & \underline{53.1}& \textbf{57.5}\\
     
					\rowcolor{gray!8} SimVG-DB~(ours) & ViT-B/32 &/& \textbf{62.1} & \underline{54.7} & \textbf{64.6} &\underline{57.2} & \textbf{54.8}& \underline{57.2}\\
					
					\hline
			\end{tabular}}
                \label{table:sotaongrec}
                \vspace{-10pt}
		\end{center}
	\end{table*}

	\begin{table*}[t]
		\setlength{\tabcolsep}{4pt}
		\renewcommand{\arraystretch}{1.1}
		\scriptsize
		\begin{center}
                \caption{Comparison with pre-trained models on RefCOCO~\cite{refcoco/+}, RefCOCO+~\cite{refcoco/+}, and RefCOCOg~\cite{refcocogumd} datasets. We only count the parameters of transformer architecture in head.}
                \vspace{-5pt}
			\begin{tabular}{l|c|c|c|ccc|ccc|cc|c}
				\hline
				\multirow{2}{*}{Models} & \multirow{2}{*}{\makecell{Visual \\ Encoder}} & Params & \multirow{2}{*}{\makecell{Pre-train \\ images}} & \multicolumn{3}{c|}{RefCOCO} & \multicolumn{3}{c|}{RefCOCO+} & \multicolumn{2}{c|}{RefCOCOg} & Time \\
				& & (M) & & val & testA & testB & val & testA & testB & val-u & test-u & (ms)\\
				
				\hline
				$\text{UNITER}_{L}$~\cite{UNITER} & RN101 & - & 4.6M & 81.41 & 87.04 & 74.17 & 75.90 & 81.45 & 66.70 & 74.86 & 75.77 & -  \\
				$\text{VILLA}_{L}$~\cite{villa} & RN101 & - & 4.6M & 82.39 & 87.48 & 74.84 & 76.17 & 81.54 & 66.84 & 76.18 & 76.71 & - \\
				MDETR~\cite{mdetr} & RN101 & 17.36 & 200K & 86.75 & 89.58 & 81.41 & 79.52 & 84.09 & 70.62 & 81.64 & 80.89 & 108 \\
				RefTR~\cite{reftr} & RN101 & 17.86 & \underline{100K} & 85.65 & 88.73 & 81.16 & 77.55 & 82.26 & 68.99 & 79.25 & 80.01 &  \textbf{40} \\
				SeqTR~\cite{seqtr} & DN53 & 7.90 & 174K & 87.00 & 90.15 & 83.59 & 78.69 &84.51 &71.87 & 82.69& 83.37  &  50 \\
				UniTAB~\cite{unitab} &RN101 &-& 200K&  88.59 & 91.06 & 83.75 & 80.97 & 85.36 &71.55& 84.58 &84.70 &  - \\
				DQ-DETR~\cite{dqdetr} & RN101 & - & 200K & 88.63 & 91.04 & 83.51 & 81.66 &86.15 &73.21 & 82.76& 83.44 &  - \\
				GroundingDINO~\cite{groundingdino} & Swin-T & - & 200K & 89.19 &  91.86 & 85.99 & 81.09 &87.40 &74.71 & 84.15& 84.94& 120 \\
				PolyFormer~\cite{polyformer} & Swin-B & - & 174K & 89.73 &  91.73 & 86.03 & 83.73 &88.60 &76.38 & 84.46& 84.96&  - \\
                PolyFormer~\cite{polyformer} & Swin-L & - & 174K & 90.38 &  92.89 & 87.16 & 84.98 &89.77 &77.97 & 85.83& 85.91 & -\\
                OFA-L~\cite{ofa} & RN152 & - & 20M & 90.05 &  92.93 & 85.26 & 85.80 &89.87 &79.22 & 85.89& 86.55 & - \\
                mPLUG-2~\cite{mplug-2} & ViT-L/14 & - & 14M & 92.40 &  94.51 & 88.42 & 86.02 &90.17 &78.17 & 85.88& 86.42 & - \\
				\hline
				\rowcolor{gray!8} SimVG-DB~(ours) & ViT-B/32 & \underline{6.32} & \textbf{28K} & {90.98} & 92.68 &{87.94} &{84.17}  &{88.58}  &{78.53} & {85.90}& 86.23& 52\\
				\rowcolor{gray!8} SimVG-TB~(ours) & ViT-B/32 & \textbf{1.58} & 174K & 90.59 & {92.80}&87.04&83.54  &88.05  &77.50&85.38& {86.28}& \underline{44}\\
				\rowcolor{gray!8} SimVG-DB~(ours) & ViT-B/32 & \underline{6.32} & 174K & {91.47} & {93.65}  & {87.94 }& {84.83} & {88.85} &{79.12} & {86.30}& {87.26}& 52 \\
                \hline
                    \rowcolor{gray!15} SimVG-TB~(ours) & ViT-L/32 & \textbf{1.58} & \textbf{28K} & \textbf{92.99} & \textbf{94.86}  & \underline{90.12}& \textbf{87.43} & \underline{91.02} &\underline{82.10} & \underline{87.95}& \underline{88.96}& 101 \\
                    \rowcolor{gray!15} SimVG-DB~(ours) & ViT-L/32 & \underline{6.32} & \textbf{28K} & \underline{92.93} & \underline{94.70}  & \textbf{90.28}& \underline{87.28} & \textbf{91.64} &\textbf{82.41} & \textbf{87.99}& \textbf{89.15}& 116 \\
				\hline
			\end{tabular}
                \label{table:sotaonpretrain}
                \vspace{-10pt}
		\end{center}
	\end{table*}

	\subsection{Ablation Studies}
	
		\begin{figure}
		\centering
            \vspace{-5pt}
		\begin{minipage}[b]{0.5\textwidth}
			\vspace{-5pt}
			\renewcommand{\arraystretch}{1.1}
			\setlength{\tabcolsep}{4pt}
			\centering
			\resizebox{\textwidth}{!}{
				\begin{tabular}{l|c|c|c}
					\hline
					\multirow{2}{*}{Method (ViT-B/32)}	& \multicolumn{3}{c}{RefCOCO} \\ \cline{2-4}
					& val & testA & testB\\
					\hline
				CLIP~\cite{CLIP}  & 73.93& 77.14&67.43 \\
					ViLT~\cite{VILT} &78.54&82.31&72.47  \\
					\rowcolor{gray!10} BEiT-3~\cite{beit3}&  82.35& 84.66 & 78.38\\
					\hline
					Baseline (BEiT-3)&  82.35& 84.66 & 78.38 \\
					+VE Interp.& 85.37(\textbf{+3.02}) & 86.67(\textbf{+2.01}) & 81.57(\textbf{+3.19}) \\
					\hline
					\rowcolor{gray!10} Token Branch&  85.47 & 86.75 & 81.66 \\
					\rowcolor{gray!10} Decoder Branch&  86.78 & 88.19  & 82.83 \\
					\hline
			\end{tabular}}
			\vspace{0pt}
			\caption{
				Some ablation experiments on different multimodal fusion architectures. VE Interp. refers to the downsampling convolution kernel in Visual Embed that performs bilinear interpolation from pre-trained weights.
			}
			\label{table:ablation-on-mixarchitecture}
			\vspace{0pt}
		\end{minipage}
		\hspace{4pt}
		\begin{minipage}[b]{0.46\textwidth}
			\setlength{\fboxrule}{0pt}
			\centering
			\fbox{\includegraphics[width=0.96\textwidth]{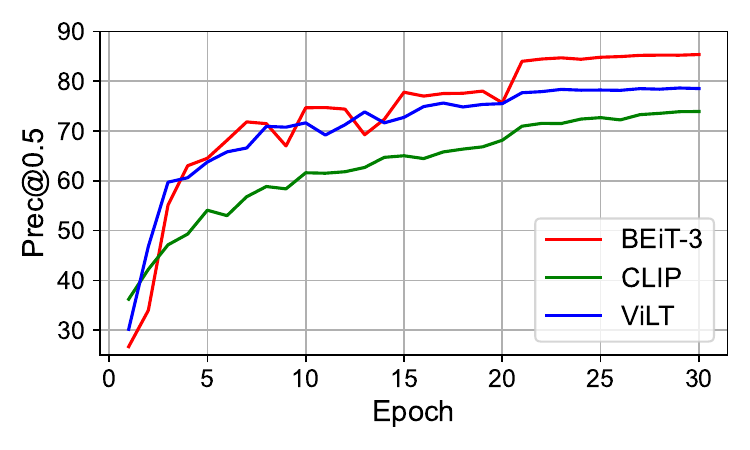}}
			\vspace{-10pt}
			\caption{
				The convergence speed of three different multimodal pretraining architecture models.
			}
			\label{fig_convergence_speed}
		\end{minipage}
		\vspace{-10pt}
	\end{figure}
	
	\subsubsection{Multi-Modality Encoder Architecture}
	{
		To investigate the advantages of decoupling multimodal fusion from visual grounding, we design three architectures for experimental verification. To ensure fairness, we consistently employ the ViT-B/32 model for feature extraction and the VGTR~\cite{vgtr} head for prediction.
        "CLIP" represents a typical dual-stream multimodal pretraining structure.
        "ViLT" represents a one-stream multimodal fusion method.
        "BEiT-3" represents a dual-stream method with a fusion encoder.
		The experimental results are reported in Table~\ref{table:ablation-on-mixarchitecture}. Approaches like ViLT and BEiT-3, which decouple the multimodal fusion process from downstream, show significant improvements compared to the methods that adopt a multimodal independent encoder architecture.
		
		However, this experiment does not aim to demonstrate the superiority of architectures like BEiT-3. 
        Our focus is to highlight that, by decoupling multimodal fusion and leveraging readily available multimodal pretrained weights, we can significantly enhance the convergence speed and performance of visual grounding. As depicted in Fig.~\ref{fig_convergence_speed}, ViLT and BEiT-3 demonstrate notably accelerated convergence by decoupling multimodal fusion. In contrast, although CLIP leverages a large amount of image-text data for pre-training, it only performs cross-modal alignment and does not integrate the information from the image and text models to achieve a fused representation.
		
        As shown in Table~\ref{table:ablation-on-mixarchitecture}, building upon BEiT-3. We observe that increasing the stride size of the original visual embedding from 16 to 32 and applying bilinear interpolation to the convolutional kernel significantly enhances performance. This is because bilinear interpolation preserves the original feature distribution after compression, thereby accelerating convergence. Furthermore, experimental results from the decoder and token branches reveal a notable performance gap, highlighting the necessity of designing dynamic weight-balance distillation to mitigate this disparity.
	}

	
	\begin{figure}
		\centering
            \vspace{-20pt}
		\begin{minipage}[b]{0.48\textwidth}
			\renewcommand{\arraystretch}{1.1}
			\setlength{\tabcolsep}{4pt}
			\centering
			\resizebox{\textwidth}{!}{
				\begin{tabular}{l|c|c|c}
					\hline
					\multirow{2}{*}{Method}	& \multicolumn{3}{c}{RefCOCO}\\ \cline{2-4}
					& val & testA & testB \\
					\hline
					\multicolumn{4}{l}{\textbf{Token Branch}} \\
					\hline
					Baseline & 85.47 & 86.75 & 81.66\\
					\rowcolor{gray!10} TQG &  86.20(\textbf{+0.73}) & 88.11(\textbf{+1.36})& 82.43(\textbf{+0.77})\\
					\hline
					\multicolumn{4}{l}{\textbf{Decoder Branch}} \\
					\hline
					Baseline & 86.78 & 88.19  & 82.83\\ 
					Mask Max Pool &  87.21&88.20	&83.28\\
					\rowcolor{gray!10} TQG & 87.44(\textbf{+0.66})& 88.84(\textbf{+0.65})& 83.61(\textbf{+0.78}) \\
					\hline
			\end{tabular}}
			\vspace{10pt}
			\caption{
				Ablation studies related to TQG module in the token and decoder branches.
			}
			\label{table:ablation-on-tgqg}
			\vspace{0pt}
		\end{minipage}
		\hspace{6pt}
		\begin{minipage}[b]{0.48\textwidth}
			\setlength{\fboxrule}{0pt}
			\centering
			\fbox{\includegraphics[width=0.96\textwidth]{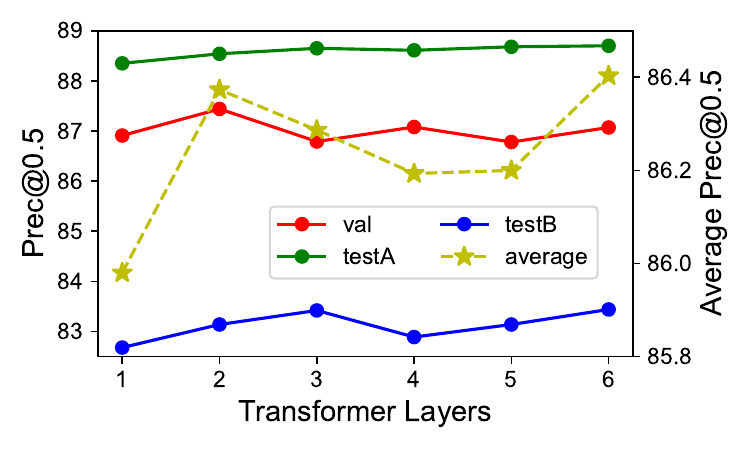}}
			\vspace{-22pt}
			\caption{
				The impact of TQG transformer layers, average refers to the mean value of three sets.
			}
			\label{fig_ablate_tgqg_layers}
		\end{minipage}
		\vspace{-10pt}
	\end{figure}

	\subsubsection{Text-guided Query Generation}
	\label{ab:tqg}
	{
		As indicated in Table~\ref{table:ablation-on-tgqg}, experimental results demonstrate a clear positive impact of the TQG module on both the token and decoder branches, achieving an average absolute improvement of 0.8 points. This guidance mechanism aligns with the concept of DAB-DETR~\cite{dabdetr}, which injects textual priors into queries to imbue them with target-pointing properties.
		"Mask Max Pool" involves using a text mask to select valid text tokens and then performing max pooling to compress the dimensions, as represented in Sec.~\ref{sect:Text-guided Query Generation}.
		Furthermore, Fig.~\ref{fig_ablate_tgqg_layers} illustrates the impact of transformer layers on TQG. The 2-layer transformer structure is adopted to balance both efficiency and performance.
	}

		\begin{figure}
		\centering
		\begin{minipage}[b]{0.46\textwidth}
			\vspace{-15pt}
			\renewcommand{\arraystretch}{1.1}
			\setlength{\tabcolsep}{4pt}
			\centering
			\resizebox{\textwidth}{!}{
				\begin{tabular}{l|c|c|c}
					\hline
					\multirow{2}{*}{Method}	& \multicolumn{3}{c}{RefCOCO} \\ \cline{2-4}
					& val & testA & testB\\
					\hline
					Baseline & 85.47 & 86.75 & 81.66 \\
					\hline
					\multicolumn{4}{l}{\textbf{One Stage Distill}} \\
					\hline
					DETR Distill\cite{chang2023detrdistill} & 86.14& 87.50& 81.54\\
					Merge Distill & 85.98& 87.27&82.09 \\
					\rowcolor{gray!10} DWB Distill & 86.57(\textbf{+1.10}) & 87.80(\textbf{+1.05}) & 82.71(\textbf{+1.05})\\
					\hline
					\multicolumn{4}{l}{\textbf{Two Stage Distill}} \\
					\hline
					DETR Distill\cite{chang2023detrdistill} &  86.49 & 88.25&82.30 \\
					Merge Distill & 86.02 & 88.03 & 82.56\\
					\rowcolor{gray!10} DWB Distill & 86.96(\textbf{+1.49}) & 88.22(\textbf{+1.47}) & 83.16(\textbf{+1.50}) \\
					\hline
			\end{tabular}}
			\vspace{0pt}
			\caption{
				Ablation studies related to DWBD module, including one/two stage distillation results.
			}
			\label{table:ablation-on-tokendistillsetting}
			\vspace{-2pt}
		\end{minipage}
		\hspace{4pt}
		\begin{minipage}[b]{0.50\textwidth}
			\setlength{\fboxrule}{0pt}
			\centering
			\fbox{\includegraphics[width=0.96\textwidth]{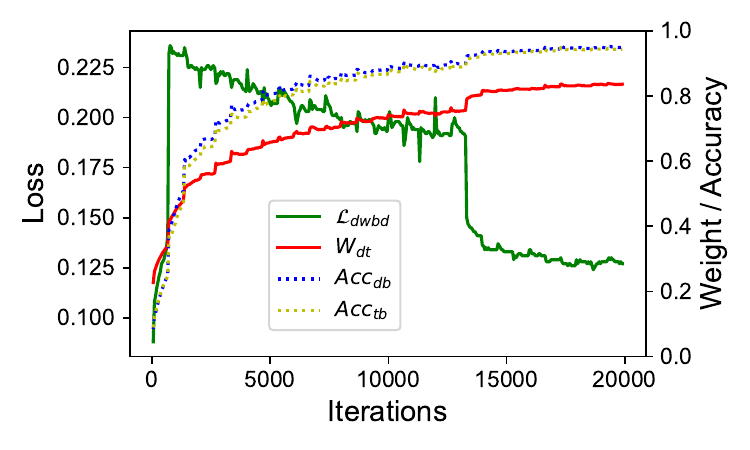}}
			\vspace{-22pt}
			\caption{
    Curves of \({\cal L}_{dwbd}\) and \(W_{dt}\), along with the accuracy of the decoder and token branches.
			}
                \vspace{-2pt}
			\label{fig_wdistill_kdloss}
		\end{minipage}
		\vspace{-10pt}
	\end{figure}

	\subsubsection{Dynamic Weight-Balance Distillation}
	\label{ab:distill}
	{
        The distillation experiments are shown in Table~\ref{table:ablation-on-tokendistillsetting}, where "DETR Distill" adopts the settings from~\cite{chang2023detrdistill} and uses the predictions from the decoder branch as the teacher for learning.
        "Merge Distill" combines the ground truth with the decoder prediction, enabling the token branch to select matching targets adaptively.
        It can be observed that all the three distillation methods improve the performance of the token branch, with the two-stage distillation method further enhancing its performance. Ultimately, our proposed DWBD achieves an average improvement of 1.5 points compared to the baseline.
        From Table~\ref{table:sotaonrec} and Table~\ref{table:sotaonpretrain}, we observe that when employing the ViT-L as teacher model, the performance of the lightweight token branch can even surpass that of the decoder branch on certain metrics during synchronous learning. We hypothesize that this is primarily because the token branch distills more robust feature representations as the teacher's cognitive capabilities improve.
        Additionally, Fig.~\ref{fig_wdistill_kdloss} illustrates the dynamic balance process of DWBD during training. We can observe that as the confidence of the decoder branch increases, the value of \(W_{dt}\) rises correspondingly, indicating that the decoder branch provides more guidance to the token branch. This mechanism allows for the dynamic adjustment of guidance distribution between the ground truth and the decoder prediction.
	}

\section{Visualization}

\begin{figure*}[t]
	\centering
	\includegraphics[width=1.0\textwidth]{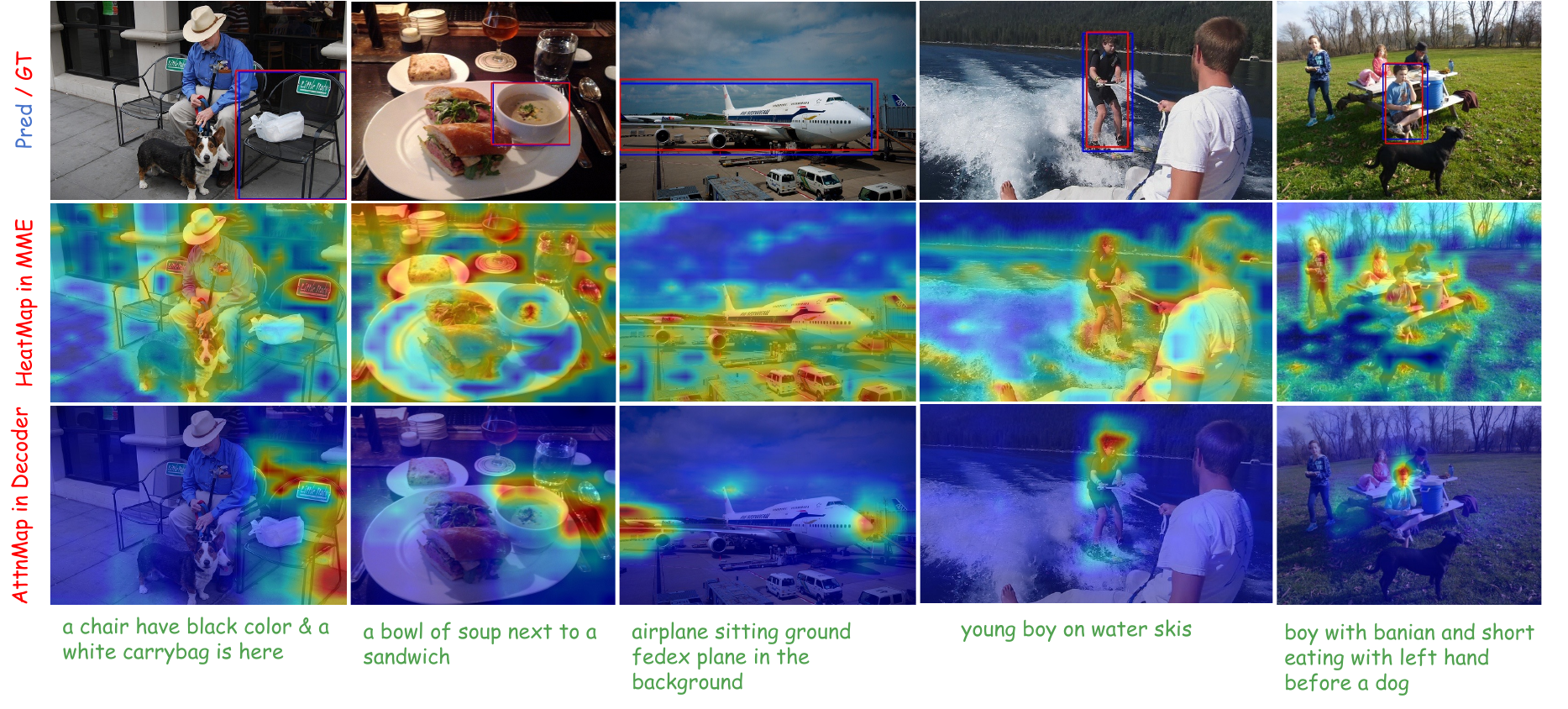}
	\caption{
			Visualization of the feature maps in MME and the decoder. The heatmap in MME is generated using GradCAM~\cite{gradcam}, while the attention response of the decoder is obtained from the attention map.
	}
	\label{fig:heatmap}
\end{figure*}

We conduct an attention analysis of SimVG from two perspectives, as shown in Fig.~\ref{fig:heatmap}. First, we visualize the multimodal representations of BEiT-3 using GradCAM~\cite{gradcam} to generate heatmaps, revealing that BEiT-3 primarily focuses on global foreground information. Additionally, we visualize the attention maps of the decoder, which highlight the model's focus on regions referred to by the text. More qualitative results can be found in Appendix~\ref{appendix:qualitative_result}.

\section{Conclusion}
In this paper, we re-examine the visual grounding task by decoupling image-text mutual understanding from the downstream task. We construct a simple yet powerful model architecture named SimVG, which leverages the existing research in multimodal fusion to fully explore the contextual associations between modalities. Additionally, to simplify the whole pipeline while maintaining performance, we adopt dynamic weight-balance distillation (DWBD) to let the stronger decoder branch guide the lightweight token branch while learning synchronously. Furthermore, we propose a text-guided query generation (TQG) module to provide textual prior knowledge for object queries. Experimental results demonstrate that SimVG not only achieves improvements in efficiency and convergence speed but also attains new state-of-the-art performance across various benchmarks.

\section*{Acknowledgements}
\label{sec:acknowledgements}
This work was supported by the National Natural Science Foundation of China under Grants 62276061.

	\medskip

	
	\bibliographystyle{plain}
	\bibliography{SimVG}
	
	\clearpage

\begin{appendices}
\section*{Appendix}
        \section{Datasets}
	\label{sec:datasets}
	\noindent{\bf RefCOCO/+/g.} 
	RefCOCO~\cite{refcoco/+} contains 142,210 referring expressions, 50,000 referred objects, and 19,994 images. The testA set primarily describes people, while the testB set mainly describes objects other than people. 
	Similarly, RefCOCO+~\cite{refcoco/+} contains 141,564 expressions, 49,856 referred objects, and 19,992 images. RefCOCO+ referring expressions focus more on attributes of the referent, such as color, shape, and digits, and avoid using words indicating absolute spatial location.
	RefCOCOg~\cite{refcocoggoogle,refcocogumd} is divided into two partitions: the \emph{google} split~\cite{refcocoggoogle} and the \emph{umd} split~\cite{refcocogumd}. Each split includes 95,010 referring expressions, 49,822 referred objects, and 25,799 images. 
	
	\noindent{\bf ReferItGame.}
	ReferItGame~\cite{referitgame} comprises 120,072 referring expressions and 99,220 referents corresponding to 19,997 images sourced from the SAIAPR-12~\cite{saiapr12} dataset. The dataset is partitioned using the cleaned Berkeley split, with 54,127, 5,842, and 60,103 referring expressions allocated to the train, validation, and test sets, respectively.
	
	\noindent{\bf Flickr30K.} 
	Flickr30K Entities~\cite{flickrentities} is characterized by short region phrases used as language queries, rather than complete sentences, which may describe multiple objects. The dataset consists of 31,783 images with a total of 427,000 referred entities across the train, validation, and test sets.
	
	\noindent{\bf GRefCOCO.} 
	GRefCOCO~\cite{GRES} comprises 278,232 expressions, which include 80,022 multi-target expressions and 32,202 no-target expressions, referring to 60,287 distinct instances in 19,994 images. Some single-target expressions are inherited from RefCOCO~\cite{refcoco/+}.
	
	\noindent{\bf Pre-Training Dataset.} 
	Following the approach in~\cite{seqtr}, we combine region descriptions from the Visual Genome~\cite{vg} dataset, annotations from RefCOCO/+/g~\cite{refcoco/+, refcocoggoogle,refcocogumd}, ReferItGame~\cite{referitgame}, and language queries from the Flickr30K Entities~\cite{flickrentities}. Our pre-training is configured in two modes. One uses only the lightweight COCO series dataset~\cite{refcoco/+, refcocoggoogle, refcocogumd}, which includes approximately 321k distinct language expressions and 28k images. The other mode is configured to be consistent with SeqTR~\cite{seqtr}, containing approximately 6.1M distinct language expressions and 174k images.

\section{Evaluation Metrics}
	\label{sect:evaluation metrics}
	\noindent{\bf Precision@0.5 [\textbf{Prec@0.5}]:}
	{
		REC and phrase localization, we evaluate the performance using Precision@0.5. The prediction is deemed correct if its IoU with ground-truth box is larger than 0.5. 
	}
	
	\noindent{\bf Precision@(F$_1$=1, IoU$\ge$0.5) [Prec@(F$_1$@0.5)]:}
	{
		The percentage of samples achieving an F$_1$ score of 1 with an IoU threshold of 0.5 is computed. A predicted bounding box is considered a TP if it has a matching ground-truth bounding box with an IoU$\geq$0.5. If multiple predicted bounding boxes match one ground-truth bounding box, only the one with the highest IoU is considered TP, and the others are FP. Ground-truth bounding boxes with no matched bounding box are FN, while predicted bounding boxes with no matched ground-truth bounding box are FP. The F$_1$ score for a sample is calculated as $F_1=\frac{2TP}{2TP+FN+FP}$. A sample is considered successfully predicted if its F$_1$ score is 1. For samples with no target, the F$_1$ score is 1 if there is no predicted bounding box, otherwise 0. The ratio of successfully predicted samples is then computed as Precision@(F$_1$=1, IoU$\ge$0.5).
	}
	
	\noindent{\bf N-acc:}
	{
		No-target accuracy (N-acc) evaluates the model's ability to identify samples with no target. In a no-target sample, predicting no bounding box is a TP, otherwise it's a FN. N-acc is calculated as $\frac{\mathit{TP}}{\mathit{TP}+\mathit{FN}}$, reflecting the model's performance in identifying samples with no target.
	}

        \section{More Implementation Details}
        \label{appendix_implementation_details}

        In this section, we provide additional details about the experimental settings described in the main text. For all ablation experiments, we use 512$\times$512 sized images as input. For ViT-B experiments, the two-stage distillation experiments use an additional 20 epochs of training. The description in Sec.~\ref{implementation details} applies to all SimVG-base models; however, we make some adjustments for the large models. 
        Due to the higher memory usage of the large models, all large models are trained with a batch size of 4. For the large model, the decoder's projection input dimension is increased from 768 to 1024, while all other settings remain consistent with the base model. Additionally, in the pre-training experiments presented in Table~\ref{table:sotaonpretrain}, the number of training epochs for the large models is reduced from 30 to 20 due to the increased training cost and the number of token branch distillation epochs is reduced from 20 to 10.
        In the experiments, all results for SimVG-TB are obtained using DWBD in a two-stage distillation process. For SimVG-DB, the results are obtained by supervising the decoder branch with ground truth.
        In the GREC experiment as shown in Table~\ref{table:sotaongrec}, we set the number of object queries to 10.
        Furthermore, the distillation parameters for all base models are set to $\gamma_1=2$ and $\gamma_2=1$, while for all large models, they are set to $\gamma_1=1$ and $\gamma_2=0.4$. All training is performed without using the exponential moving average (EMA) strategy. 
        Lastly, it is important to emphasize that the BEiT-3 pre-trained model used in this paper was not trained on the six datasets used for validation in this study.

        \section{Additional Exploration Studies}

	\subsection{Number of Layers and Query}
\begin{figure}[ht]
		\centering
		\begin{minipage}[b]{0.45\textwidth}
			\setlength{\fboxrule}{0pt}
			\centering
			\includegraphics[width=0.97\textwidth]
			{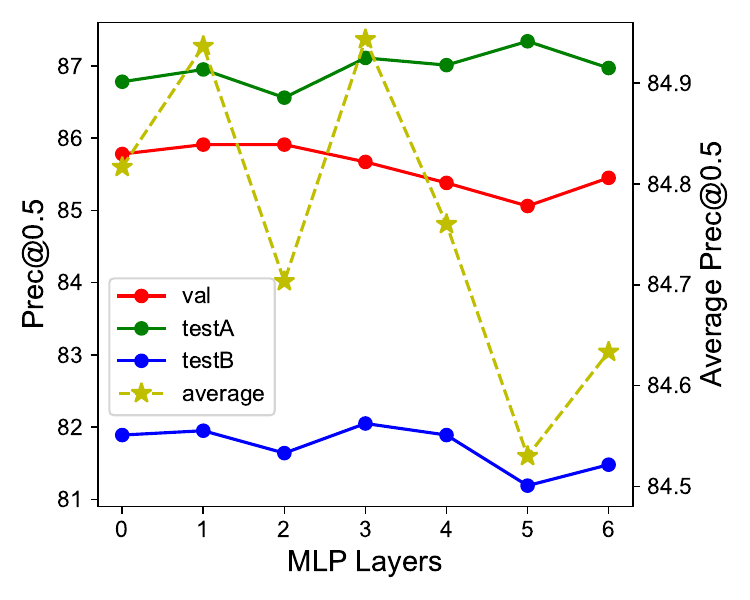}
			\vspace{-10pt}
			\caption{
				Ablation study on number of token MLP layers in RefCOCO dataset.
			}
			\label{fig_ablate_tokenlayer}
		\end{minipage}
		\hspace{5px}
		\begin{minipage}[b]{0.45\textwidth}
			\setlength{\fboxrule}{0pt}
			\centering
			\includegraphics[width=0.97\textwidth]{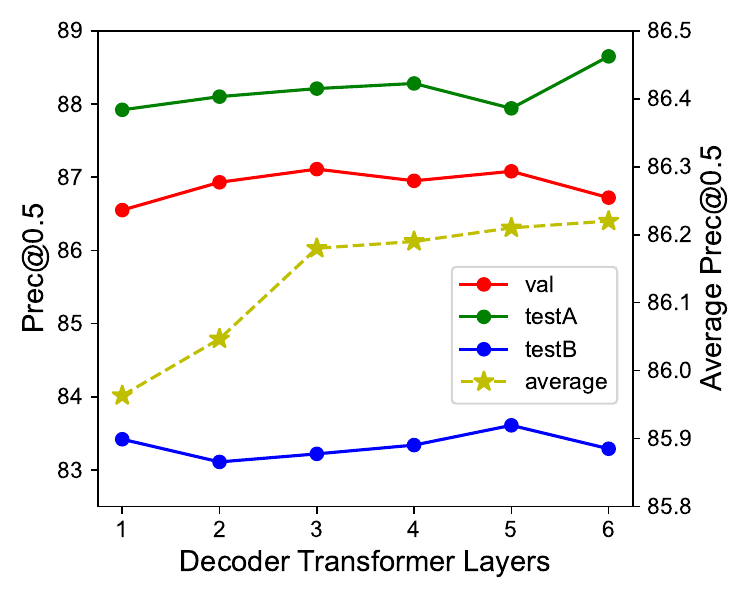}
			\captionof{figure}{
				Ablation study on number of decoder transformer layers in RefCOCO dataset.
			}
			\label{fig_ablate_decoderlayer}
		\end{minipage}
		\vspace{-10pt}
	\end{figure}

	{
		The number of transformer layers in the decoder branch and MLP layers in the token branch can impact performance and efficiency. Excessive layers can reduce computational efficiency and increase parameter count, while too few layers may lead to subpar performance. To determine the optimal number of layers, we conducted experiments under varying settings, with results presented in Fig.~\ref{fig_ablate_tokenlayer} and Fig.~\ref{fig_ablate_decoderlayer}.
		Fig.~\ref{fig_ablate_tokenlayer} indicates that increasing the number of MLP layers does not yield significant gains. Therefore, we selected single MLP layer. In contrast, Fig.~\ref{fig_ablate_decoderlayer} shows that increasing the number of transformer layers results in performance improvement. Considering efficiency, we opted for 3 transformer layers.
	}
	
        \subsection{Efficiency of Training}
        We analyze the training efficiency of SimVG from two aspects. As shown in Table~\ref{table:epoch_traintime}, SimVG significantly outperforms mainstream methods in terms of the required number of epochs and training time. On the RefCOCO+ dataset, SimVG only requires 30 epochs and 5.5 hours, faster than SeqTR's 60 epochs and 9 hours, demonstrating the accelerated convergence brought by multimodal pre-training. As shown in Table~\ref{table:params_pretrainimgs}, SimVG adopts a more lightweight head design and, despite using less pre-training data, achieves notable performance improvements.

        \begin{figure}
		\centering
		\tiny
		\begin{minipage}[b]{0.434\textwidth}
			\renewcommand{\arraystretch}{1.1}
			\setlength{\tabcolsep}{4pt}
			\centering
			\resizebox{\textwidth}{!}{
				\begin{tabular}{l|c|c|c}
					\hline
					\multirow{2}{*}{Method} & \multicolumn{1}{c|}{Visual} & \multirow{2}{*}{Epoch} & \multicolumn{1}{c}{Training} \\
                    &Encoder& &Time \\ 
     		    \hline
					TransVG & RN50 & 180 & \textasciitilde90h\\
					VGTR & RN50 & 120 & \textasciitilde50h \\
					Dynamic MDETR & ViT-B/16& 90 & -\\
					SeqTR & DN53& 60&\textasciitilde9h \\
                    \hline
					\rowcolor{gray!10} SimVG & ViT-B/32 & 30 & \textasciitilde5.5h \\	
					\hline
			\end{tabular}}
			\vspace{4pt}
			\caption{
			     Training time on a single RTX 4090 GPU and the number of epochs required for convergence on the RefCOCO+ dataset.
			}
			\label{table:epoch_traintime}
		\end{minipage}
		\hspace{5pt}
		\begin{minipage}[b]{0.53\textwidth}
			\renewcommand{\arraystretch}{1.1}
			\setlength{\tabcolsep}{4pt}
			\centering
			\resizebox{\textwidth}{!}{
				\begin{tabular}{l|c|c|c|c|c}
					\hline
					\multirow{2}{*}{Method}	& \multicolumn{1}{c|}{Head} & \multicolumn{1}{c|}{Pretrain} & \multicolumn{3}{c}{RefCOCO}\\ \cline{4-6}
					& Params (M) & Images & val & testA & testB \\
					\hline
					RefTR & 17.86 & 100K & 85.65& 88.73 & 81.16\\
					MDETR & 17.36 & 200K & 86.75 & 89.58 & 81.41\\ 
					SeqTR & 7.90 & 174K & 87.00 & 90.15 & 83.59 \\
                    \hline
                    \rowcolor{gray!10} SimVG-DB & 6.32 & 28K & 90.98 & 92.68 & 87.94 \\
                    \rowcolor{gray!10} SimVG-TB & 1.58 & 28K & 90.18 & 92.41 & 87.21 \\
					\hline
			\end{tabular}}
			\vspace{4pt}
			\caption{
				A comparison of the number of parameters in the head and the number of pre-training images for some publicly available methods.
			}
			\label{table:params_pretrainimgs}
		\end{minipage}
	\end{figure}

        \subsection{Compare with Multi-modal Large Language Models}
	{
        With the surge of large models across various domains, there have been methods proposed for the visual grounding task that leverage multimodal large language models. These methods combine autoregressive and prompt-based approaches to locate targets. Most of these models have billions of parameters, benefiting from their base pretrained models being pretrained on very large datasets, thus exhibiting strong robust performance. We compare these large model approaches with our proposed SimVG, as shown in Table~\ref{tab:mllm}. Despite our model having an order of magnitude fewer parameters compared to large models, it still achieves competitive performance, thanks to its decoupled multimodal understanding. Moreover, from the experiments, we also observe that as the model's parameter count increases, SimVG's performance shows an increasing trend.

         \begin{table*}[t]
            \centering
            \scriptsize
            \renewcommand\arraystretch{1.4}
            \begin{tabular}{lc|ccc|ccc|cc}
            \hline
            \multirow{2}{*}{Models} & \multirow{2}{*}{LLM Size} & \multicolumn{3}{c|}{RefCOCO} & \multicolumn{3}{c|}{RefCOCO+} & \multicolumn{2}{c}{RefCOCOg}\\
            \cline{3-10}  & & val & testA & testB & val & testA & testB & val & test \\
            \hline
            KOSMOS-2~\cite{kosmos}   & 1.6B  & 52.32 & 57.42 & 47.26 & 45.48 & 50.73 & 42.24 & 60.57 & 61.65\\ 
            Shikra~\cite{shikra}     & 7B  & 87.01 & 90.61 & 80.24 & 81.60 & 87.36 & 72.12 & 82.27 & 82.19\\
            NExT-Chat*~\cite{nextchat} & 7B & 85.50  & 90.00  & 77.90  & 77.20  & 84.50  & 68.00  & 80.10  & 79.80 \\
            Ferret*~\cite{ferret}    & 7B & 87.49 & 91.35 & 82.45 & 80.78 &{87.38} & 73.14 & {83.93} & {84.76}\\
            GroundingGPT~\cite{groundinggpt} & 7B & {88.02} & {91.55} & {82.47} & {81.61} & 87.18 & {73.18} & 81.67 & 81.99  \\
            PixelLLM~\cite{pixelllm} & 4B & {89.80} & {92.20} & {86.40} & {83.20} & 87.00 & {78.90} & 84.60 & 86.00 \\
            COMM~\cite{comm} & 7B & \underline{91.73} & \underline{94.06} & \underline{88.85} & \underline{87.21} & \textbf{91.74} & \underline{81.39} & \underline{87.32} & \underline{88.33} \\
            \hline
            \rowcolor{gray!10}SimVG-DB-Base~(ours) & \textbf{0.18B} & {91.47} & {93.65}  & {87.94 }& {84.83} & {88.85} &{79.12} & {86.30}& {87.26}\\
            \rowcolor{gray!10}SimVG-DB-Large~(ours) & \underline{0.61B} &\textbf{92.87} & \textbf{94.35}  & \textbf{89.46}& \textbf{87.28} & \underline{91.64} &\textbf{82.41} & \textbf{87.99}& \textbf{89.15} \\
            \hline  
            \end{tabular}
            \caption{Performance comparison on the REC task. "*" indicates that the model employs additional image region perception modules.}
            \label{tab:mllm}
        \end{table*}
	}

        \begin{wrapfigure}{r}{0.4\textwidth}
			\vspace{-50pt}
			\centering
			\includegraphics[width=0.38\textwidth]{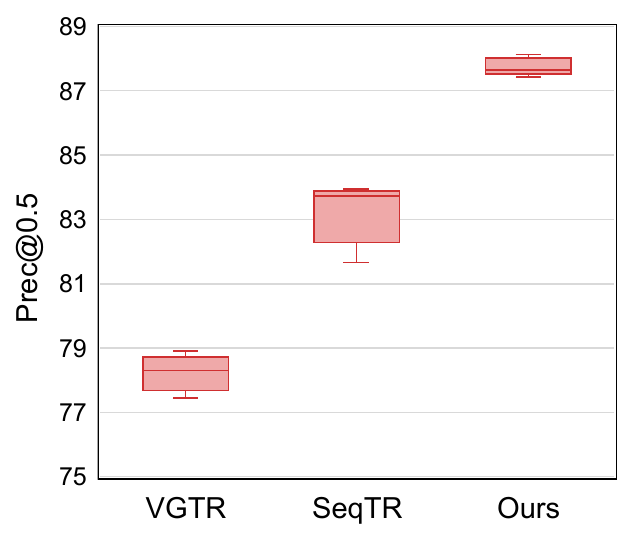}
			\vspace{-10pt}
			\caption{
				The box-plot on validation set of RefCOCO dataset.
			}
			\label{fig_error_bar}
			\vspace{-30pt}
		\end{wrapfigure}

		\subsection{Analysis of Error Bars}
		\label{errorbar}
		We conducted experiments with VGTR~\cite{vgtr}, SeqTR~\cite{seqtr}, and our proposed SimVG, each repeated five times with different random seeds. As illustrated in Fig.~\ref{fig_error_bar}, our approach exhibits a notably higher median accuracy compared to the VGTR~\cite{vgtr} method. Furthermore, in contrast to SeqTR~\cite{seqtr}, our method demonstrates a more tightly clustered distribution of results. These results indicate that the proposed SimVG not only enhances accuracy but also increases stability.

        \section{Limitations}
        \label{limitation}
        
        Our method does not fully explore or utilize hierarchical information in features. Approaches like the FPN in ViTDet~\cite{vitdet} that expand feature hierarchy could be considered to further enhance the model's ability to capture targets of different scales. Our method can be applied not only to detection-related tasks but also to segmentation-related tasks. Further validation on more downstream tasks is warranted to demonstrate its stable effectiveness.

        \section{Broader Impacts}
        \label{Broader Impacts}
        Further research and careful consideration are necessary when utilizing this technology, as the presented proposed method relies on statistics derived from training datasets that may possess biases and could potentially result in negative societal impacts.

	\section{Qualitative Results}
	\label{appendix:qualitative_result}
	In this section, we present the visualizations of the SimVG results for the referring expression comprehension~(REC) and general referring expression comprehension task~(GREC). We present both the caption and image. The red and blue rectangles on images refer to groundtruth and predict boxes, respectively.

        \begin{figure*}[t]
		\centering
		\includegraphics[width=0.8\textwidth]{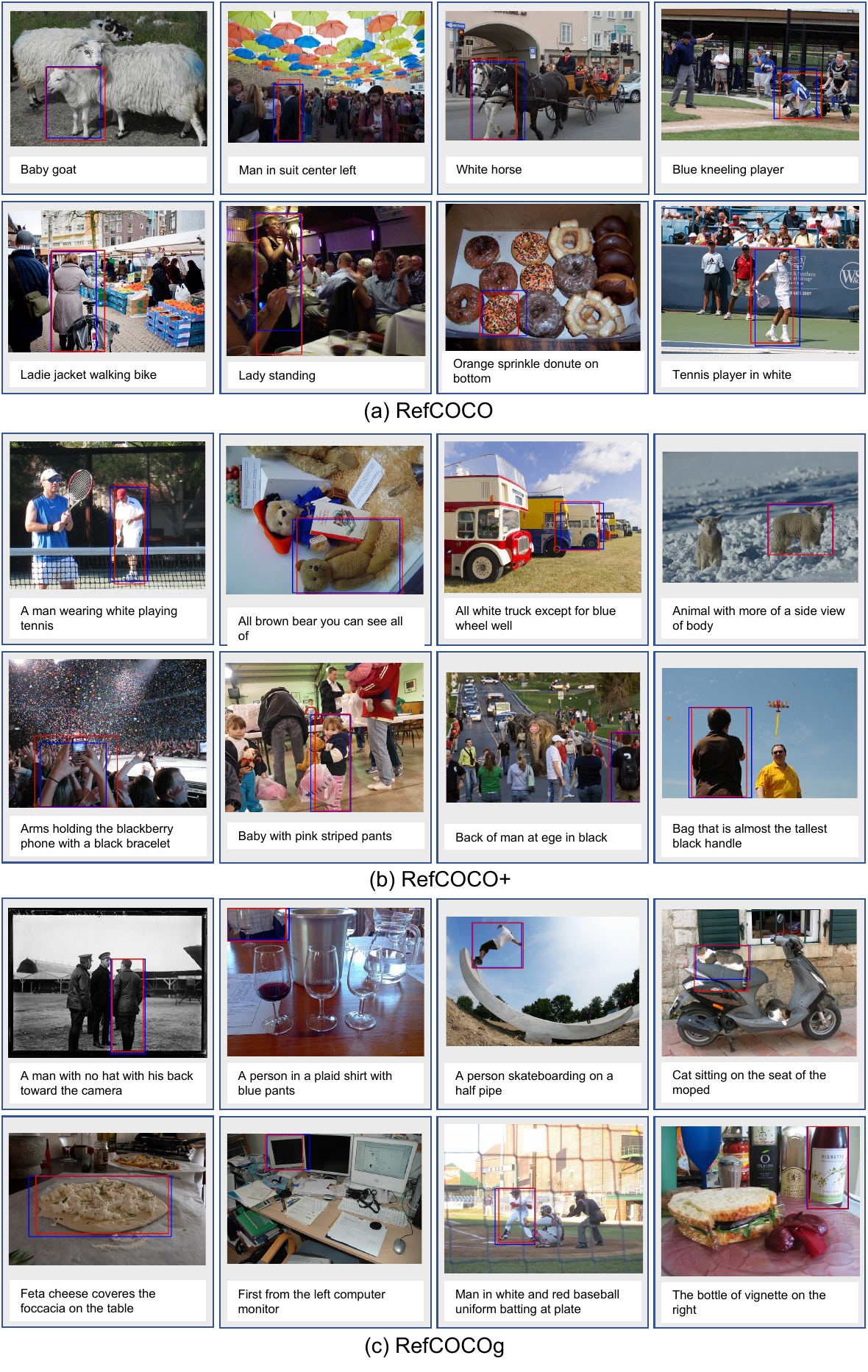}
		\caption{
			Examples predicted by SimVG on the validation set of RefCOCO/+/g datasets.
		}
		\label{fig:visualization}
	\end{figure*}

        Firstly, in Fig.~\ref{fig:visualization}, we present some visual examples of our proposed SimVG on the RefCOCO, RefCOCO+, and RefCOCOg datasets. It can be seen that SimVG can accurately perceive and locate objects even in long texts or complex images.

        \begin{figure*}[t]
		\centering
		\includegraphics[width=0.8\textwidth]{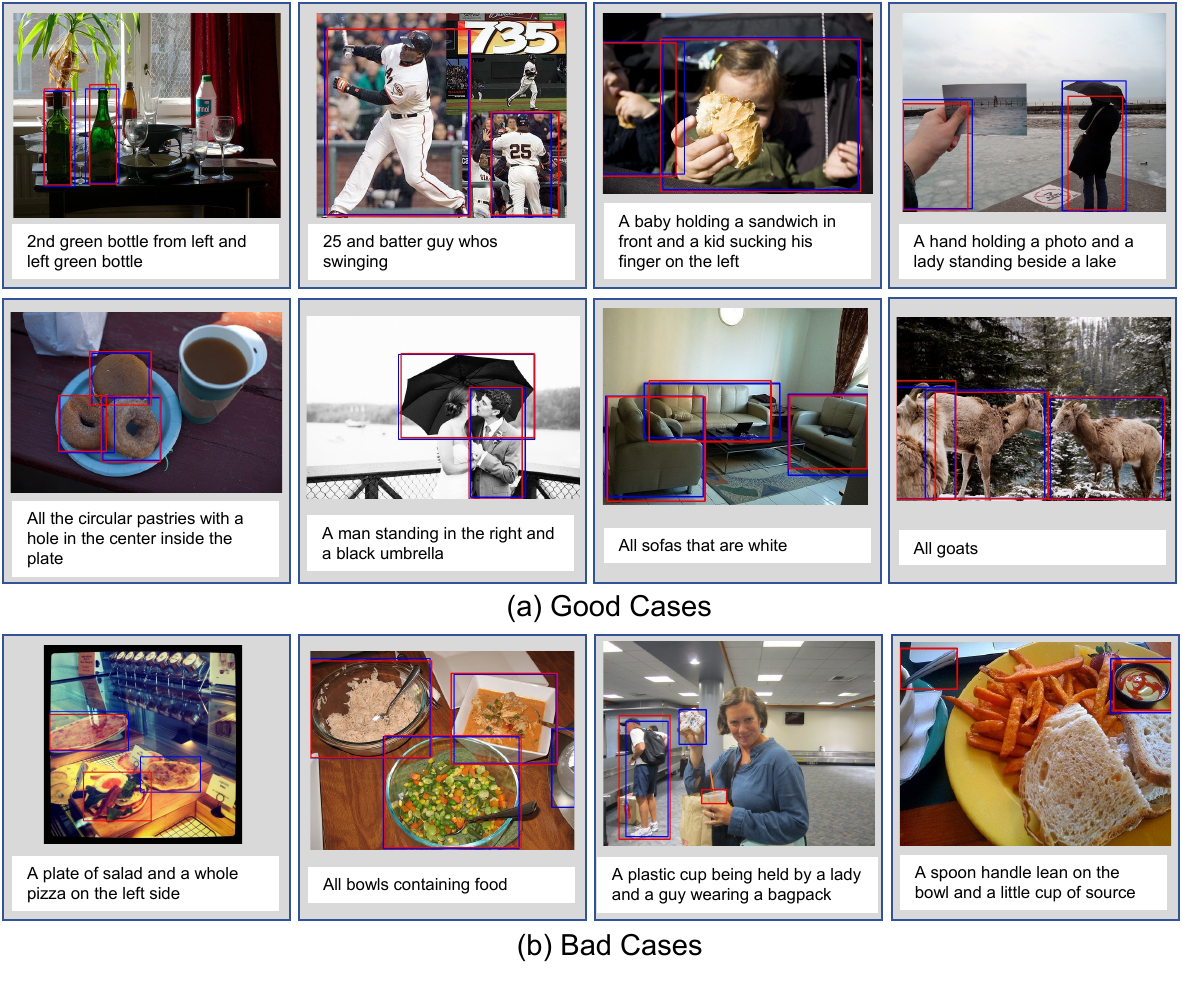}
		\caption{
			Examples predicted by SimVG on the validation set of GRefCOCO datasets.
		}
		\label{fig:visualization_grec}
	\end{figure*}
        \begin{figure*}[t]
		\centering
        \includegraphics[width=0.8\textwidth]{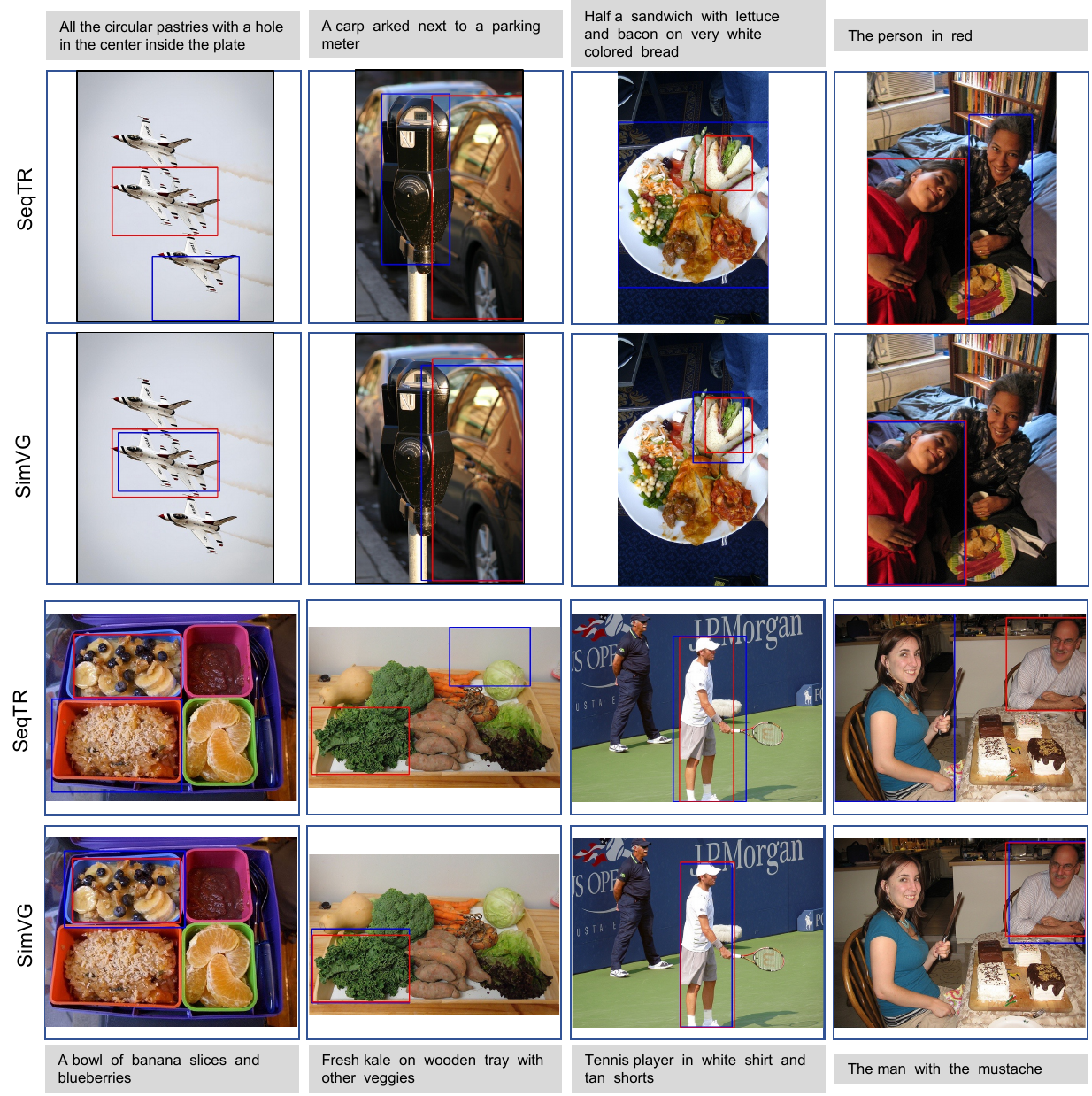}
		\caption{
			Some examples comparing SeqTR and SimVG models on the RefCOCOg dataset.
		}
		\label{fig:visualization_compare}
	\end{figure*}

        Then, in Fig.~\ref{fig:visualization_grec}, we show some examples of correct and incorrect results of SimVG on the GRefCOCO dataset. The GREC task requires a deeper semantic understanding of image-text relationships, and in the error cases, the instances of missed detections and false positives significantly increase.

        Finally, in Fig.~\ref{fig:visualization_compare}, we compare the output results of SimVG with the comparable SeqTR model. We find that the proposed SimVG model can better understand the interrelationships between images and texts.

\end{appendices}

\clearpage

\newpage

\newpage
\section*{Checklist}
 
\begin{enumerate}

\item {\bf Claims}
    \item[] Question: Do the main claims made in the abstract and introduction accurately reflect the paper's contributions and scope?
    \item[] Answer: \answerYes{} 
    \item[] Justification: We list the main contributions of this paper at the end of the introduction, and explain the scrope and differences between our method and previous methods in both the abstract and the introduction.

\item {\bf Limitations}
    \item[] Question: Does the paper discuss the limitations of the work performed by the authors?
    \item[] Answer: \answerYes{} 
    \item[] Justification: We describe the limitations of our approach in Sec.~\ref{limitation}.

\item {\bf Theory Assumptions and Proofs}
    \item[] Question: For each theoretical result, does the paper provide the full set of assumptions and a complete (and correct) proof?
    \item[] Answer: \answerNA{} 
    \item[] Justification: We present comprehensive experimental results conducted across multiple datasets, comparing our method with prior works to validate its effectiveness.

    \item {\bf Experimental Result Reproducibility}
    \item[] Question: Does the paper fully disclose all the information needed to reproduce the main experimental results of the paper to the extent that it affects the main claims and/or conclusions of the paper (regardless of whether the code and data are provided or not)?
    \item[] Answer: \answerYes{} 
    \item[] Justification: We describe the structure of our proposed model throughout Sec.~\ref{methods} and implementation details of the experiment in Sec.~\ref{implementation details}. To further ensure reproducibility, we will soon open source the code of our method.

\item {\bf Open access to data and code}
    \item[] Question: Does the paper provide open access to the data and code, with sufficient instructions to faithfully reproduce the main experimental results, as described in supplemental material?
    \item[] Answer: \answerYes{} 
    \item[] Justification: Once the paper is accepted, we will open source our code. In addition, the data we use are all publicly available.

\item {\bf Experimental Setting/Details}
    \item[] Question: Does the paper specify all the training and test details (e.g., data splits, hyperparameters, how they were chosen, type of optimizer, etc.) necessary to understand the results?
    \item[] Answer: \answerYes{} 
    \item[] Justification: Implementation Details in Sec.~\ref{implementation details} and Appendix Sec.~\ref{appendix_implementation_details}.

\item {\bf Experiment Statistical Significance}
    \item[] Question: Does the paper report error bars suitably and correctly defined or other appropriate information about the statistical significance of the experiments?
    \item[] Answer: \answerYes{} 
    \item[] Justification: We independently set up repeated experiments to analyze the error bar in Sec.~\ref{errorbar}.

\item {\bf Experiments Compute Resources}
    \item[] Question: For each experiment, does the paper provide sufficient information on the computer resources (type of compute workers, memory, time of execution) needed to reproduce the experiments?
    \item[] Answer: \answerYes{} 
    \item[] Justification: Implementation Details in Sec.~\ref{implementation details} and Appendix Sec.~\ref{appendix_implementation_details}.
    
\item {\bf Code Of Ethics}
    \item[] Question: Does the research conducted in the paper conform, in every respect, with the NeurIPS Code of Ethics \url{https://neurips.cc/public/EthicsGuidelines}?
    \item[] Answer: \answerYes{} 
    \item[] Justification: We have carefully read the NeurIPS Code of Ethics and have determined that our approach adheres to the relevant ethical guidelines.

\item {\bf Broader Impacts}
    \item[] Question: Does the paper discuss both potential positive societal impacts and negative societal impacts of the work performed?
    \item[] Answer: \answerYes{} 
    \item[] Justification: Although our work is only for academic research purpose, we also discuss the potential positive societal impacts and negative societal impacts in Sec. \ref{Broader Impacts}.
    
\item {\bf Safeguards}
    \item[] Question: Does the paper describe safeguards that have been put in place for responsible release of data or models that have a high risk for misuse (e.g., pretrained language models, image generators, or scraped datasets)?
    \item[] Answer: \answerNA{} 
    \item[] Justification: Our work poses no such risks.

\item {\bf Licenses for existing assets}
    \item[] Question: Are the creators or original owners of assets (e.g., code, data, models), used in the paper, properly credited and are the license and terms of use explicitly mentioned and properly respected?
    \item[] Answer: \answerYes{} 
    \item[] Justification:
    
    1. RefCOCO:https://paperswithcode.com/dataset/refcoco
    
    2. GRefCOCO:https://paperswithcode.com/dataset/grefcoco
    
    3. ReferItGame:https://paperswithcode.com/dataset/referitgame
    
    4. Flickr30K Entities:https://paperswithcode.com/dataset/flickr30k-entities

\item {\bf New Assets}
    \item[] Question: Are new assets introduced in the paper well documented and is the documentation provided alongside the assets?
    \item[] Answer: \answerNA{} 
    \item[] Justification: Our work does not release new assets. The data and models used in our work are publicly released.

\item {\bf Crowdsourcing and Research with Human Subjects}
    \item[] Question: For crowdsourcing experiments and research with human subjects, does the paper include the full text of instructions given to participants and screenshots, if applicable, as well as details about compensation (if any)? 
    \item[] Answer: \answerNA{} 
    \item[] Justification: Our work does not involve human subjects.

\item {\bf Institutional Review Board (IRB) Approvals or Equivalent for Research with Human Subjects}
    \item[] Question: Does the paper describe potential risks incurred by study participants, whether such risks were disclosed to the subjects, and whether Institutional Review Board (IRB) approvals (or an equivalent approval/review based on the requirements of your country or institution) were obtained?
    \item[] Answer: \answerNA{} 
    \item[] Justification: Our work does not involve human subjects.

\end{enumerate}
\end{document}